\documentclass[10pt]{article} 
\usepackage[accepted]{rlc}

\usepackage{amssymb}            
\usepackage{mathtools}          
\usepackage{mathrsfs}           
\usepackage{graphicx}           
\usepackage{subcaption}         
\usepackage[space]{grffile}     
\usepackage{url}                
\hypersetup{
 colorlinks=True,
 linkcolor=nhred,
 citecolor=gray,
 urlcolor=nhred}
\usepackage{wrapfig}
\usepackage{amsmath}
\usepackage{amsthm}
\usepackage{amsfonts}
\usepackage{nicefrac}
\usepackage{enumitem}
\usepackage{colortbl}
\usepackage{algorithm}
\usepackage{algpseudocode} 
\usepackage{bm} 
\usepackage[normalem]{ulem}
\usepackage[format=plain,
            labelfont=it,
            labelsep=period]{caption}
\usepackage{booktabs}
\usepackage{bbding}
\usepackage{fancyvrb}
\usepackage{listings}
\usepackage{multirow}

\definecolor{codegreen}{rgb}{0,0.5,0}
\definecolor{codered}{rgb}{0.7,0.1,0.1}
\definecolor{codegray}{rgb}{0.5,0.5,0.5}
\definecolor{codepurple}{rgb}{0.58,0,0.82}
\definecolor{backcolour}{rgb}{1,1,1}
\lstdefinestyle{python}{
    language=Python,
    backgroundcolor=\color{backcolour},   
    commentstyle=\color{codered}\textit,
    keywordstyle=\bfseries\color{codegreen},
    numberstyle=\tiny\color{codegray},
    stringstyle=\color{codepurple},
    basicstyle=\ttfamily\scriptsize,
    breakatwhitespace=false,         
    breaklines=true,                 
    captionpos=b,                    
    keepspaces=true,                 
    numbers=left,                    
    numbersep=4pt,                  
    showspaces=false,                
    showstringspaces=false,
    showtabs=false,                  
    tabsize=1,
    fancyvrb=true
}
\definecolor{nhred}{HTML}{D62727}
\definecolor{nhteal}{HTML}{66C2A4}
\definecolor{cella}{rgb}{1.0, 0.92, 0.92}

\definecolor{citecolor}{HTML}{0071bc}
\definecolor{BrickRed}{HTML}{AA4A44}

\title{A Recipe for Unbounded Data Augmentation in \\Visual Reinforcement Learning}

\author{Abdulaziz Almuzairee\thanks{Sponsored by an institutional fellowship from Kuwait University. The support is gratefully acknowledged.} \\
    aalmuzairee@ucsd.edu \\
    UC San Diego
    \And
    Nicklas Hansen \\
    nihansen@ucsd.edu\\
    UC San Diego
    \And
    Henrik I. Christensen \\
    hichristensen@ucsd.edu\\
    UC San Diego}

\begin{document}

\maketitle

\begin{abstract}
$Q$-learning algorithms are appealing for real-world applications due to their data-efficiency, but they are very prone to overfitting and training instabilities when trained from visual observations. Prior work, namely SVEA, finds that selective application of data augmentation can improve the visual generalization of RL agents without destabilizing training. We revisit its recipe for data augmentation, and find an assumption that limits its effectiveness to augmentations of a photometric nature. Addressing these limitations, we propose a generalized recipe, SADA, that works with wider varieties of augmentations. We benchmark its effectiveness on DMC-GB2 -- our proposed extension of the popular DMControl Generalization Benchmark -- as well as tasks from Meta-World and the Distracting Control Suite, and find that our method, SADA, greatly improves training stability and generalization of RL agents across a diverse set of augmentations. \\ \\
\textbf{Visualizations, code and benchmark: \href{https://aalmuzairee.github.io/SADA}{https://aalmuzairee.github.io/SADA}}

\end{abstract}

\section{Introduction}
\label{sec:introduction}

Visual Reinforcement Learning (RL) has a myriad of real-world applications \citep{mnih2013playing, levine2016end, pinto2016supersizing, kalashnikov2018qt, Berner2019Dota2W, Vinyals2019GrandmasterLI}, and visual $Q$-learning algorithms are especially enticing because of their potential for high data-efficiency. However, they are very prone to overfitting on their training distribution due to the combination of flexible representation, high-dimensional data, and limited visual diversity in training environments \citep{Peng_2018, cobbe2019quantifying, Julian2020EfficientAF}.

Data augmentation is a widely used technique for learning visual invariances in supervised learning \citep{noroozi2016unsupervised, tian2019contrastive, chen2020simple}, but has been found to cause training instabilities when applied to visual RL \citep{Lee2019ASR, laskin2020reinforcement, hansen2021softda}. Prior work, SVEA \citep{hansen2021stabilizing}, found that a more selective application of data augmentation in the critic update of actor-critic algorithms \citep{lillicrap2015continuous, haarnoja2018soft} improved training stability significantly. The actor (\emph{policy}) -- which shares its visual backbone with the critic (\emph{$Q$-function}) -- is then optimized solely from unaugmented observations. By sharing their visual backbone, the actor indirectly benefits from the learned invariances.

\begin{figure}
    \centering
    \includegraphics[width=0.9\linewidth]{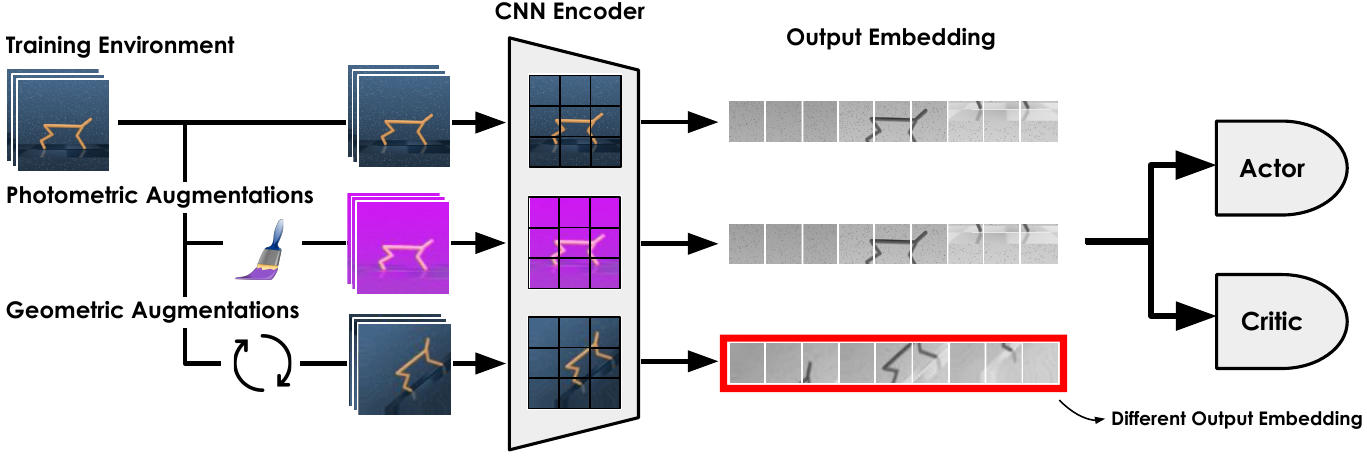}
    \caption{\textbf{Augmentation Effect on CNN Output.} We illustrate how the output embedding of a trained CNN changes wrt. image augmentations. The output of unaugmented and photometrically augmented images are identical due to the ability of a CNN to learn color invariances. However, the output of a CNN is generally not invariant to \emph{geometric} augmentations (\emph{e.g.}, rotation).}
    \label{fig:main_fig}
    \vspace{-0.2in}
\end{figure}

In this work, we revisit the data augmentation recipe proposed in SVEA, and discover an \emph{assumption} that limits its practicality to augmentations of a photometric (color or light altering) nature. SVEA \emph{assumes} that an encoder's output embedding can become fully invariant to input augmentations. If an encoder's output is fully invariant to input augmentations, then an actor, only trained on unaugmented observations, can become robust to input augmentations indirectly through a shared actor-critic encoder. However, this leads to two key failure cases: \emph{(i)} the output of a convolutional neural network (CNN) encoder can not become invariant to input \emph{geometric} augmentations \emph{e.g.}, rotation or translation; (see Figure~\ref{fig:main_fig}) and \emph{(ii)} the encoder and critic are trained end-to-end, thus, part of the invariance may be off-loaded to the critic regardless of the type of augmentation.

To address these limitations, we propose \textbf{SADA}: \textbf{S}tabilized \textbf{A}ctor-Critic under \textbf{D}ata \textbf{A}ugmentation, a generalized data augmentation recipe that supports a wide variety of augmentations. Instead of only augmenting critic inputs, SADA augments \emph{both} actor and critic inputs, but does so carefully to avoid training instabilities: \emph{(1)} in actor updates, only the policy input is augmented while the $Q$-function input is left unaugmented, \emph{(2)} in critic updates, only the online $Q$-function input is augmented while the target $Q$-function input is unaugmented, and \emph{(3)} we jointly optimize components on both augmented and unaugmented data. Importantly, SADA requires no additional forward passes, losses, or parameters.

To stress-test our method, we propose DMC-GB2, an extension of the DeepMind Control Suite Generalization Benchmark \citep{hansen2021softda} that encompasses a wider and more challenging collection of test environments than existing benchmarks. We benchmark methods across DMC-GB2, tasks from Meta-World \citep{yu2020meta}, and the Distracting Control Suite \citep{stone2021distracting}, and find that SADA greatly improves training stability and generalization of RL agents under a diverse set of augmentations.

\vspace{-0.08in}
\section{Prior Work on Data Augmentation for Visual RL}
\label{sec:related}
The practice of learning visual invariances by augmenting data is ubiquitous in machine learning literature, and has been studied extensively in the context of supervised and self-supervised learning algorithms for computer vision problems \citep{noroozi2016unsupervised, wu2018unsupervised, oord2018representation, tian2019contrastive, chen2020simple, he2022masked}. More recently, use of augmentation has also been popularized in the context of visual RL. However, there is mounting evidence that much of the wisdom and practices developed in other areas (\emph{e.g.} computer vision) do not translate to RL problems, presumably due to differences in learning objectives, datasets, and network architectures used. For example, while machine learning literature commonly considers a \emph{fixed} dataset, RL algorithms are often trained on a non-stationary data distribution (replay buffer) that changes throughout training, and incoming data is typically a function of the current (behavioral) policy. As a result, RL datasets are often small and have limited diversity. This section provides an overview of prior work that leverages data augmentation to improve \emph{data-efficiency} and \emph{generalization}.

\textbf{Improving \emph{data-efficiency} with data augmentation.} Much of the existing literature on visual RL leverages \emph{weak} data augmentation (\emph{e.g.} random crop or image shift) as a regularizer when data is limited, \emph{i.e.}, when data-efficiency is critical \citep{srinivas2020curl, laskin2020reinforcement, kostrikov2020image, stooke2020atc, yarats2021reinforcement, Hansen2022pre}, without particular emphasis on generalization or robustness to changes in the environment. For example, seminal works RAD \citep{laskin2020reinforcement} and DrQ \citep{kostrikov2020image} demonstrate that randomly cropping or shifting images, respectively, by a few pixels greatly improves data-efficiency and training stability of $Q$-learning algorithms -- even when agents are trained and tested in the same environment. However, \citet{laskin2020reinforcement} simultaneously find that other types of augmentation (rotation, random convolution, masking) lead to training instabilities and a substantial \emph{decrease} in data-efficiency.

\textbf{Improving \emph{generalization} with data augmentation.} Visual generalization is a challenging but increasingly important problem in RL due to its limited data diversity. Multiple prior works aim to improve the training stability and generalization of RL algorithms by, \emph{e.g.}, proposing new types of augmentation \citep{Lee2019ASR, wang2020improving, hansen2021softda, hansen2021stabilizing, zhang2021generalization, wang2023generalizable}, or introducing new (auxiliary) objectives \citep{Raileanu2020AutomaticDA, hansen2021deployment, Wang2021UnsupervisedVA, fan2021secant, yuan2022don, yang2024movie}. For example, \citet{Lee2019ASR} augment high-frequency content in observations using random convolution, \citet{hansen2021softda} randomly overlay observations with out-of-domain images, and \citet{yang2024movie} adapt to camera changes at test-time using an auxiliary self-supervised objective and augmented data. Perhaps most importantly, SVEA \citep{hansen2021stabilizing} investigate \emph{why} strong augmentations (such as those used in the aforementioned works) often destabilize training in an RL context, and propose an alternative method of applying augmentations that mitigate these instabilities. Our work builds upon SVEA and demonstrates that -- while SVEA is robust to \emph{photometric} augmentations -- it largely fails when applied to (equally important) \emph{geometric} augmentations.

We recommend the survey by \citet{kirk2023survey} for a more comprehensive overview of prior work.

\vspace{-0.08in}
\section{Background \& Definitions}
\label{sec:prelim}
\textbf{Visual Reinforcement Learning} (RL) formulates interaction between an agent and its environment as a Partially Observable Markov Decision Process (POMDP) \citep{kaelbling1998pomdp}. A POMDP can be formalized as a tuple $(\mathcal{S}, \mathcal{O}, \mathcal{A}, \mathcal{T}, R, \gamma)$, where $\mathcal{S}$ is an unobservable state space, $\mathbf{o} \in \mathcal{O}$ are observations from the environment (\emph{e.g.}, RGB images), $\mathbf{a} \in \mathcal{A}$ are actions, $\mathcal{T \colon \mathcal{S} \times \mathcal{A} \mapsto \mathcal{S}}$ is a transition function, $r$ is a task reward from a reward function  $R \colon \mathcal{S} \times \mathcal{A} \mapsto \mathbb{R}$, and $\gamma$ is a discount factor. Throughout this work, we approximate the unobservable states $\mathbf{s} \in \mathcal{S}$ by defining observations as a stack of the three most recent RGB frames $\mathbf{o}_{t} \doteq \{ \mathbf{x}_{t}, \mathbf{x}_{t-1}, \mathbf{x}_{t-2} \}$ for frames $\mathbf{x}_{t:t-2}$ at time $t$ \citep{mnih2013playing}. The goal is then to learn a policy $\pi \colon \mathcal{O} \mapsto \mathcal{A}$ such that the discounted sum of rewards $\mathbb{E}_{\pi} \left[ \sum_{t=0}^{\infty} \gamma^{t} r_{t} \right]$ is maximized (in expectation) when following the policy $\pi$.

\textbf{$Q$-Learning} algorithms developed for visual RL generally estimate the optimal state-action value function $Q^{*} \colon \mathcal{O} \times \mathcal{A} \mapsto \mathbb{R}$ with a neural network (denoted the \emph{critic}). This is achieved by dynamic programming using the single-step Bellman error $Q(\mathbf{o}_{t}, \mathbf{a}_{t}) - y_{t}$ where $y_{t}$ is the temporal difference (TD) target $y_{t} \doteq r_{t} + \gamma Q(\mathbf{o}_{t+1}, \mathbf{a}_{t+1}),~\mathbf{a}_{t+1} \sim \pi(\cdot | \mathbf{o}_{t+1})$. In practice, the $Q$-network used to compute $y_{t}$ is usually chosen to be an exponential moving average of the $Q$-function being learned \citep{lillicrap2015continuous, haarnoja2018soft}. The policy $\pi$ is obtained by taking the action $\mathbf{a}_{t} \approx \arg\max_{\mathbf{a}_{t}} Q(\mathbf{o}_{t}, \mathbf{a}_{t}) ~\forall \mathbf{o}_{t}$ in the current dataset (replay buffer), which is typically estimated by training a separate \emph{actor} network when $\mathcal{A}$ is continuous. These two components -- the actor and the critic -- are iteratively updated by collecting data in the environment, appending it to a replay buffer $\mathcal{D}$, and optimizing $Q,\pi$ with the following objectives using stochastic gradient descent:
\begin{align}
    \label{eq:actor-critic}
    \mathcal{L}_{Q}(\mathcal{D}) = \mathbb{E}_{(\mathbf{o}_{t},\mathbf{a}_{t}, r_{t}, \mathbf{o}_{t+1})\sim\mathcal{D}} &\left[ \| Q(\mathbf{o}_{t}, \mathbf{a}_{t}) - y_{t} \|_{2} \right]~~~\textcolor{gray}{\textrm{(critic)}}\\
    \label{eq:actor}
    \mathcal{L}_{\pi}(\mathcal{D}) = \mathbb{E}_{\mathbf{o}_{t} \sim \mathcal{D}} &\left[ -Q(\mathbf{o}_{t}, \pi(\mathbf{o}_{t}) \right]~~~~~~~~\textcolor{gray}{\textrm{(actor)}}
\end{align}
where gradients of the first objective are computed wrt. $Q$ only, and gradients of the second objective are computed wrt. $\pi$ only. When learning from images, observations are commonly encoded using a shared convolutional encoder $f$ such that $Q,\pi$ are redefined as $Q(f(\mathbf{o}_{t}), \mathbf{a})$ and $\pi(f(\mathbf{o}_{t}))$, with $f$ only being updated by the critic objective $\mathcal{L}_{Q}$. Due to the recurrent and self-referential nature of Equations \ref{eq:actor-critic}-\ref{eq:actor}, $Q$-learning algorithms are often more data-efficient than other algorithm classes, but are very prone to training instabilities -- especially when data augmentation is applied to observations.

\textbf{Image transformations.} Throughout this work, we classify image transformations into two types: \emph{photometric} and \emph{geometric} transformations. Photometric transformations alter image color and lighting properties while preserving the spatial arrangement of pixels (\emph{e.g.} random convolution, image overlay). Geometric transformations alter the spatial arrangement of pixels while keeping image color and lighting properties intact (\emph{e.g.} rotation, shift). We visualize examples of photometric and geometric transformations in Figure \ref{fig:main_fig}.

\vspace{-0.08in}
\section{Stabilized Actor-Critic Learning under Data Augmentation}
\label{sec:method}

 We revisit common wisdom and practices when applying data augmentation in $Q$-learning algorithms, and discover that prior work makes an assumption that only holds for augmentations that are photometric in nature.  We propose \textbf{SADA}: \textbf{S}tabilized \textbf{A}ctor-Critic Learning under \textbf{D}ata \textbf{A}ugmentation, a generalized recipe for data augmentation that significantly improves the performance of a wider variety of augmentations. We start by outlining the assumptions of prior work, its implications, and then present our proposed solution.

\begin{figure}
    \centering
    \includegraphics[width=1.0\linewidth]{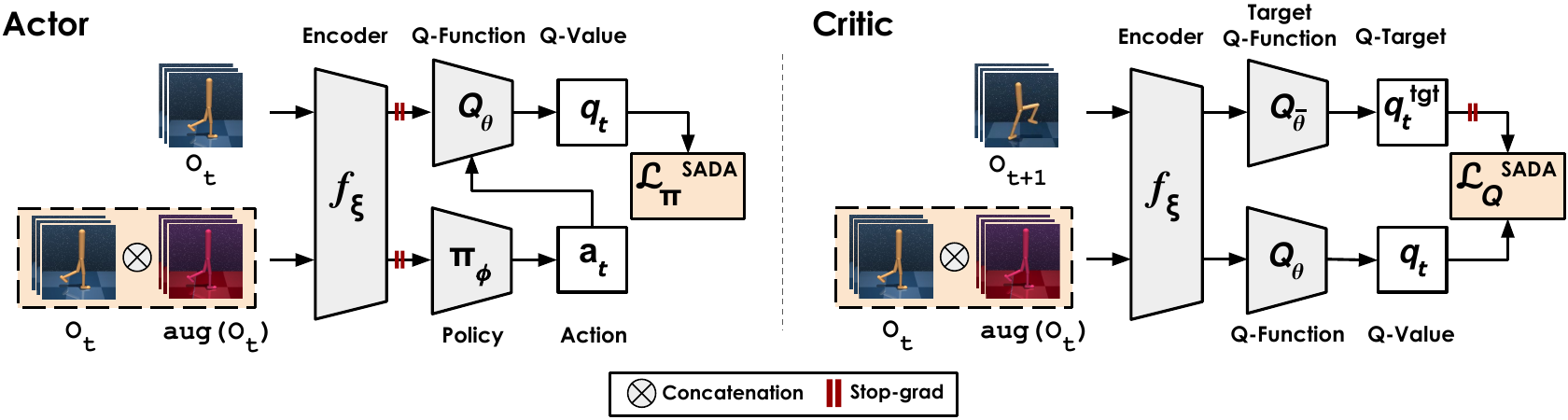}
    \vspace{-0.25in}
    \caption{\textbf{Our approach.} Overview of SADA applied to a generic actor-critic algorithm. We highlight our algorithmic contributions in yellow. SADA \emph{selectively} applies augmentations to the actor and critic inputs, and modifies the learning objectives accordingly.}
    \label{fig:model_diag}
    \vspace{-0.15in}
\end{figure}

\vspace{-0.07in}
\subsection{Shortcomings of Prior Work}
\label{sec:shortcomings}

Naive augmentation, where all inputs are indiscriminately augmented, has been shown to lead policies to suboptimal convergence \citep{Raileanu2020AutomaticDA, hansen2021stabilizing}. Unlike supervised learning, the application of augmentation in RL can lead to a conflict of task objective, conflict of learning objective, or increased variance that exacerbates instabilities within actor-critic frameworks. 

To stabilize actor-critic learning under \emph{strong} applications of data augmentation, SVEA \citep{hansen2021stabilizing} selectively applies augmentations in the critic updates, without any application of augmentation in the actor updates. The actor -- optimized purely from unaugmented observations -- becomes robust to augmentations indirectly, through the use of a shared actor-critic encoder. By using this formulation, SVEA \emph{assumes} that the encoder can output embeddings that are invariant to input augmentations, such that an actor can indirectly become robust to input augmentations. This assumption leads to two key failure cases: \emph{(i)} the output embedding of a CNN encoder can not become invariant to input \emph{geometric} augmentations, \emph{(ii)} even with \emph{photometric} augmentations, part of the robustness could be off-loaded to the critic. 

We provide a motivating example for key failure case (i) in Figure \ref{fig:main_fig} and show that geometric transformations will always induce changes in a CNN's output embedding. Therefore, an actor not directly trained on these changed output embeddings will not become robust to these geometric transformations. As for key failure case (ii), a CNN can learn to output embeddings that are invariant to input photometric augmentations. However, the objective is formulated such that the output of the critic is robust to input image augmentations, indicating that if either the encoder or the critic is robust, the objective will be satisfied. Therefore, some of the photometric resistance might be contained within the critic, rendering the actor weaker against photometric transformations.

\vspace{-0.07in}
\subsection{Our Proposed Recipe}
\label{sec:approach}

To mitigate shortcomings of previous works, the actor needs to \emph{directly} train on the augmented stream. However, naively training the actor on the augmented stream exacerbates training instabilities. Each image augmentation applied adds a more complex distribution for the agent to learn compared to the original training distribution. To overcome this complexity, we introduce SADA, a general framework for stabilizing actor-critic agents under \emph{strong} applications of data augmentation. 

In the actor's update, we elect to use asymmetric observation inputs to the policy and $Q$-function \citep{pinto2017asymmetric}. Specifically, we allow the policy to observe \emph{both} the augmented and unaugmented streams, while the $Q$-function estimates the $Q$-value observing \emph{only} the unaugmented stream. Since the $Q$-value estimates of both the augmented and unaugmented streams should be identical, we allow the $Q$-function to exploit only the unaugmented stream (easier distribution) in making accurate $Q$-value estimates. Given an observation $\mathbf{o}_t$, replay buffer $\mathcal{D}$, and an encoder $f_\xi$, the actor objective for a generic actor critic thus becomes:
\begin{align}
    \mathcal{L}_{\pi_\phi}^\textbf{SADA}(\mathcal{D}) = \mathbb{E}_{\mathbf{o}_{t} \sim \mathcal{D}} &\left[ -Q_\theta(\mathbf{m}_t, \pi_\phi(\mathbf{p}_t)) \right]~~~~~~~~\textcolor{gray}{\textrm{(actor)}}
\end{align}
where $\mathbf{p}_t = f_\xi(\left[\mathbf{o}_{t}, \mathbf{o}_{t}^\text{aug}\right]_N), ~~ \mathbf{m}_t = f_\xi(\left[\mathbf{o}_{t}, \mathbf{o}_{t}\right]_N)$ and $\mathbf{o}_t^\text{aug} = \text{aug}(\mathbf{o}_t,v_t), v_t \sim \mathcal{V}$. 
We use $[\cdot]_N$ to denote concatenation for batch size of dimensionality $N$ where $\mathbf{o}_t, \mathbf{o}_t^\text{aug} \in \mathbb{R}^{N\times C\times H\times W}$. We use aug() as the augmentation operator where we stochastically sample from the augmentation distribution $\mathcal{V}$ and apply it to the input observation. 
 
 In the critic update, we apply a similar asymmetric observation setup with the $Q$-value and $Q$-target estimates. We allow the online $Q$-function, $Q_\theta$, to estimate the $Q$-value observing both the augmented and unaugmented streams, while the target $Q$-function, $Q_{\overline{\theta}}$, estimates the $Q$-targets observing only the unaugmented stream. Since the $Q$-target estimates of both the augmented and unaugmented streams should be identical,  this reduces the variance in $Q$-target estimates and allows the target $Q$-function to exploit the unaugmented stream (easier distribution) in making accurate $Q$-target estimates. The $Q$-target estimate, $q_t^{tgt}$, is unaltered while the critic objective, $\mathcal{L}^\textbf{SADA}_{Q_{\theta}}$, is changed such that:
\begin{align}
  q_t^{tgt} = r(\mathbf{o}_t, \mathbf{a_t}) + \gamma \text{max}_{\mathbf{a'_t}}&Q_{\overline{\theta}}(f_\xi(\mathbf{o}_{t+1}),\mathbf{a'}) \\
    \label{eq:sada-critic}
    \mathcal{L}^\textbf{SADA}_{Q_{\theta}}(\mathcal{D}) = \mathbb{E}_{(\mathbf{o}_{t},\mathbf{a}_{t}, r_{t}, \mathbf{o}_{t+1})\sim\mathcal{D}} &\left[ \| Q_\theta(\mathbf{p}_{t}, \mathbf{a}_{t}) - \mathbf{y}_{t} \|_{2} \right]~~~\textcolor{gray}{\textrm{(critic)}}
\end{align}
where $\mathbf{p}_t = f_\xi(\left[ \mathbf{o}_t,\mathbf{o}_t^{\text{aug}}\right]_{N})$,~ and $\mathbf{y}_t = \left[ q_t^{tgt},q_t^{tgt}\right]_{N}$. An overview of our algorithm is provided in Figure~\ref{fig:model_diag}. A detailed SAC-based formulation is provided in Appendix \ref{subsection:app_sac}, and pseudocode is provided in Appendix \ref{subsection:app_pseudo}.

\vspace{-0.08in}
\section{Experiments}
\label{sec:experiments}

We evaluate our method and baselines across 11 visual RL tasks from the DMControl \citep{tassa2018deepmind} and Meta-World-v2 \citep{yu2020meta} benchmarks and 12 test distributions from our proposed DMControl - Generalization Benchmark 2 (DMC-GB2). We additionally evaluate on the Distracting Control Suite \citep{stone2021distracting} and provide the results in Appendix \ref{subsection:app_dcs_results}. All methods are trained under \emph{strong} augmentations in the training environments and evaluated in a zero-shot manner on their respective test distributions. See Figure \ref{fig:dmc_main} for a visualization of DMC-GB2 test environments. The full DMControl and Meta-World task lists are provided in Appendix \ref{subsection:app_train_setup}. Concretely, we aim to answer the following questions through experimentation: 

\begin{figure}[t]
    \centering
    \begin{minipage}[t]{1.0\textwidth}
        \centering
        \includegraphics[width=\linewidth]{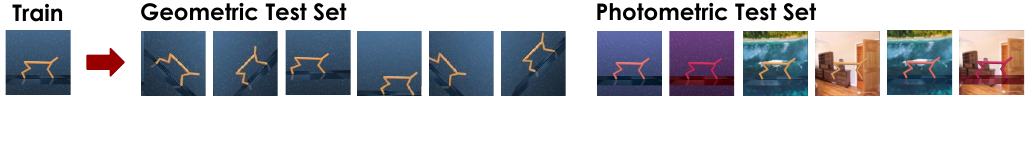}
    \end{minipage}
    \vspace{-0.625in}
    
    \begin{center}
        \textcolor{gray}{
            \rule{0.95\textwidth}{0.05cm}
        }
    \end{center}

    \begin{minipage}[t]{1.0\linewidth}
        \centering
        \includegraphics[width=0.94\linewidth,trim={0 0 0 3cm},clip]{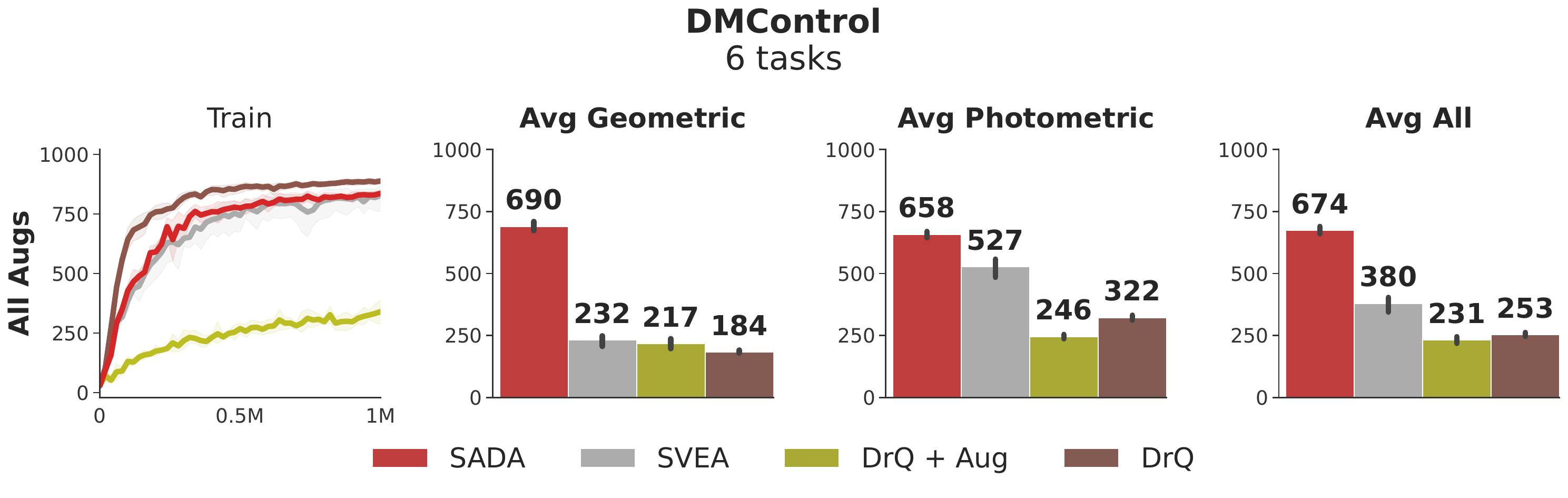}
    \end{minipage}
    \vspace{-0.225in}
    \caption{\textbf{Overall Robustness.} \emph{(Top)} Samples from the DMC-GB2 test distributions, divided into geometric and photometric test sets. \emph{(Bottom)} Episode reward on DMC-GB2 when trained under all (geometric and photometric) augmentations, averaged across all DMControl tasks. Mean and 95\% CI over 5 seeds.}
    \label{fig:dmc_main}
    \vspace{-0.15in}
\end{figure}

\vspace{-0.175in}
\begin{itemize}[label={-},leftmargin=0.25in]
    \setlength\itemsep{-0.25em}
    \item \textbf{Robustness.} How does SADA compare to baselines in terms of \emph{overall} augmentation robustness? In terms of \emph{geometric} vs \emph{photometric} robustness?
    \item \textbf{Analysis.} Why do baselines fail to display \emph{geometric} robustness, and how does SADA solve the problem? How do each of the SADA \emph{design choices} affect results?
    \item \textbf{Generality.} Can SADA be \emph{readily applied} to other RL backbones and benchmarks?
\end{itemize}
\vspace{-0.1in}

\textbf{Setup.} We build on  DrQ \citep{laskin2020reinforcement} as our backbone algorithm, and use a fixed set of hyperparameters across all tasks and environments. All agents are trained for one million environments steps and use stacks of the three most recent RGB frames ($3\times\mathbb{R}^{\mathbf{(3\times84\times84)}})$ as observations. A full list of hyperparameters and training details is provided in Appendix \ref{sec:app_back}.

\textbf{Test environments.} 
As our experiments will reveal, previous methods largely fail to generalize to geometric changes, which has gone unnoticed due to existing benchmarks predominantly evaluating photometric robustness. Therefore, we propose the Deepmind Control Suite Generalization Benchmark 2 (DMC-GB2), an extension of DMC-GB \citep{hansen2021softda} to encompass a wider collection of photometric and geometric test distributions. In DMC-GB2, we provide geometric and photometric test sets. The geometric test set considers two types of geometric distributions -- rotations and shifts -- both individually and jointly, and at varying intensities categorized as \emph{easy} and \emph{hard} environments. The photometric test set considers a complementary setup for two types of photometric distributions -- colors and videos. Detailed visualizations of all 12 DMC-GB2 test distributions is provided in Appendix \ref{subsection:app_dmc}.

\textbf{Data augmentation.} We apply a \emph{weak} augmentation of random shifting to all our inputs as conducted in our DrQ baseline, and consider it unaugmented. We further employ a set of \emph{strong} augmentations, taking into account both \emph{geometric} and \emph{photometric} transformations. For geometric augmentations, we use random shift \citep{laskin2020reinforcement}, random rotation, and a combination consisting of random rotation followed by random shift. For photometric augmentations we use random convolution \citep{Lee2019ASR}, random overlay \citep{hansen2021softda}, and a combination consisting of random convolution followed by random overlay. We sample an augmentation from this set of six \emph{strong} augmentations for each input sample in all our experiments unless stated otherwise. Detailed visualizations of all augmentations is provided in Appendix \ref{subsection:app_augs}.

\textbf{Baselines.} We benchmark our method against the following well-established baselines. 1) \textbf{DrQ} \cite{laskin2020reinforcement}, a visual based Soft Actor Critic baseline that uses random shifts as the default augmentation for all inputs.  2) \textbf{DrQ + Aug}, a variant of DrQ implemented with a naive application of \emph{strong} augmentations. 3) \textbf{SVEA} \cite{hansen2021stabilizing}, which builds on DrQ with a selective application of augmentation in the $Q$-function to increase robustness under \emph{strong} augmentations.    

\vspace{-0.07in}
\subsection{Results \& Discussion}
\label{sec:experiments-results}

\begin{figure}
    \centering
    \includegraphics[width=1.0\linewidth,trim={0 0 0 3.2cm},clip]{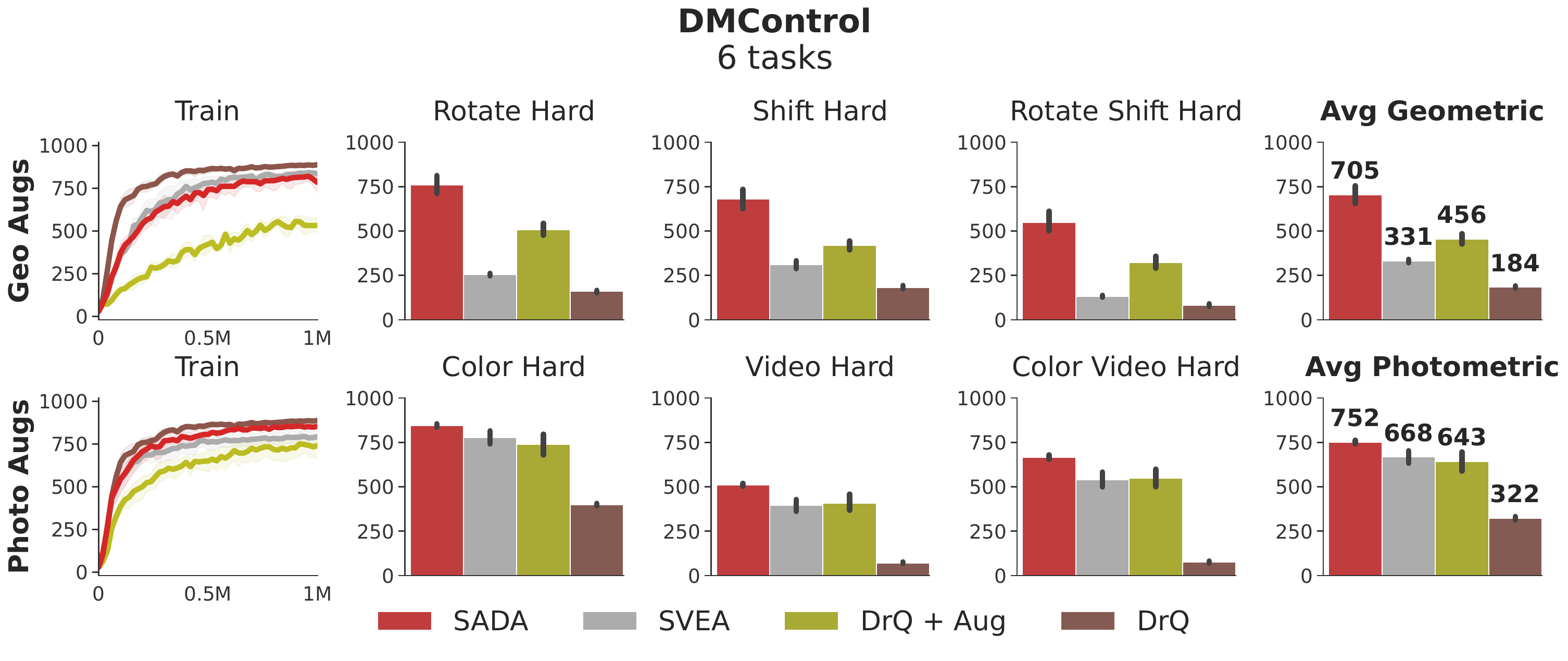}
    \vspace{-0.2in}
    \caption{\textbf{Geometric vs Photometric Robustness.} Episode reward averaged over all DMControl tasks. \emph{(Top)} Trained under geometric augmentations and evaluated on DMC-GB2 geometric test set. \emph{(Bottom)} Trained under photometric augmentations and evaluated on DMC-GB2 photometric test set. All hard levels visualized. Mean and 95\% CI over 5 random seeds.
    }
    \label{fig:exp_sep}
    \vspace{-0.1in}
\end{figure}

\textbf{Robustness.} For measuring the \emph{overall} robustness, we train all methods under all augmentations (geometric and photometric augmentations) and evaluate them on all DMC-GB2 test sets (geometric and photometric test sets).  As our empirical results indicate in Figure \ref{fig:dmc_main}, SADA's \emph{overall} robustness surpasses the baselines in all DMC-GB2 test sets by a large margin (77\%), all while attaining a similar sample efficiency to its unaugmented DrQ baseline on the training environment.

To analyze the \emph{geometric} vs \emph{photometric} robustness of each method, we conduct another experiment where we train under each set of augmentations separately. We train under strong geometric augmentations and evaluate on the geometric test set, and follow a complementary setup under strong photometric augmentations with the photometric test set. We visualize the results in Figure \ref{fig:exp_sep} along with all the individual hard level intensities in our test suite. SADA \emph{consistently} shows superior robustness, outperforming baselines in all separate test sets and individual levels while achieving similar training sample efficiency to its unaugmented DrQ baseline. Extended results and per-task breakdowns are provided in Appendix \ref{sec:app_extended_results}.

\textbf{Analysis.} While baselines show varying degrees of photometric robustness, they \emph{fail to display geometric robustness} in Figures \ref{fig:dmc_main} and \ref{fig:exp_sep}. For the DrQ baseline, geometric transformations are out of its training distribution. With naive application of data augmentation, DrQ + Aug achieves poor training sample efficiency. To achieve high training sample efficiency, SVEA selectively applies augmentation in the critic update. Nevertheless, this performance does not translate to the geometric test distributions due to key failure case (i), the output embedding of a CNN can not become invariant to input geometric transformations. This failure case can only be resolved with an actor \emph{directly} trained on the input augmentations. When the actor is \emph{directly} trained on the input augmentations using SADA's objective, the agent is able to achieve high training sample efficiency that effectively translates to the geometric test distributions. 

\begin{wrapfigure}[13]{r}{0.24\textwidth}
    \vspace{-0.2in}
    \centering
    \includegraphics[width=0.2\textwidth,trim={3.1cm 0.9cm 2.2cm 2.1cm},clip]{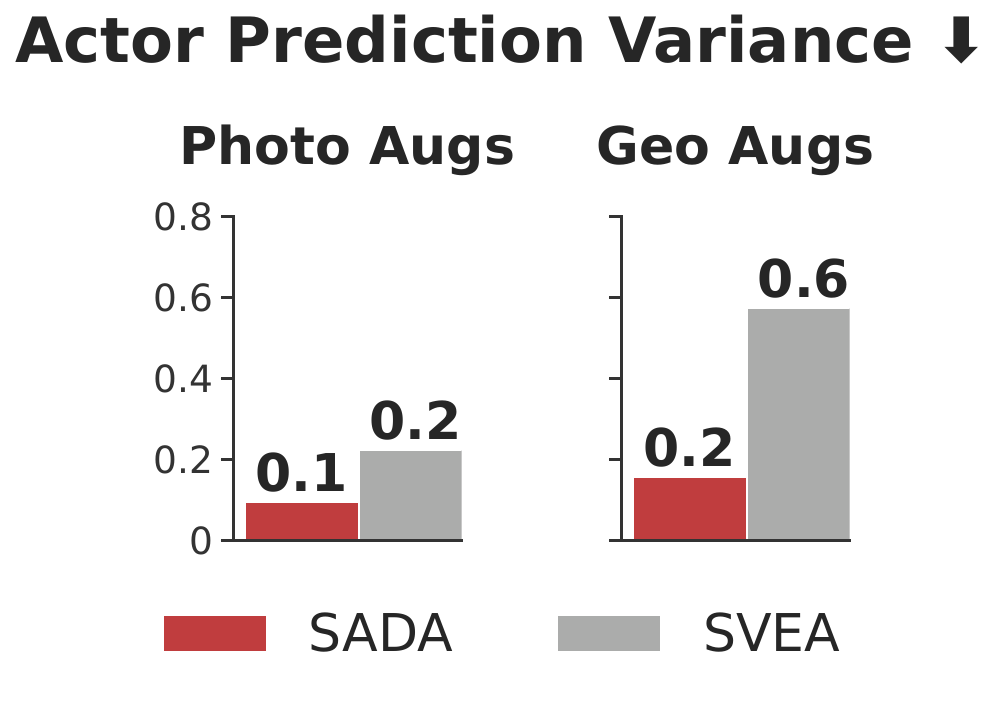}
    \caption{\textbf{Actor Prediction Variance.} Actor prediction variance between augmented and unaugmented observations. $\mathbf{\downarrow}$ Lower is better.}
    \label{fig:act_var}
\end{wrapfigure}

Even in terms of photometric robustness, SADA surpasses all baselines, including SVEA. This is mainly due to failure case (ii) of SVEA's assumption, where some of the augmentation robustness is contained within the critic and not the encoder. This can also be resolved by training the actor \emph{directly} on the input augmentations using SADA's objective.

To \emph{quantitatively} assess the augmentation robustness of converged SADA and SVEA agents, we measure the variance of actions predicted on the augmented observations with respect to the unaugmented observations in Figure \ref{fig:act_var}. Despite being trained on all augmentations, SVEA's action predictions have high variance when observing geometric augmentations as opposed to photometric augmentations, confirming SVEA's shortcomings. For a \emph{qualitative} assessment, we utilize T-SNE to visualize the encoder's output embedding before being fed into the actor and the critic in Appendix \ref{subsection:app_tsne}. Our findings reveal that photometric distributions can share the same space in the latent embedding as the original training distribution, while geometric distributions are distant in the latent space and seem to have little overlap with the training distribution, necessitating the need to \emph{directly} train the actor on them.

\begin{figure}
    \centering
    \includegraphics[width=1.0\linewidth,trim={0 0 0 3.2cm},clip]{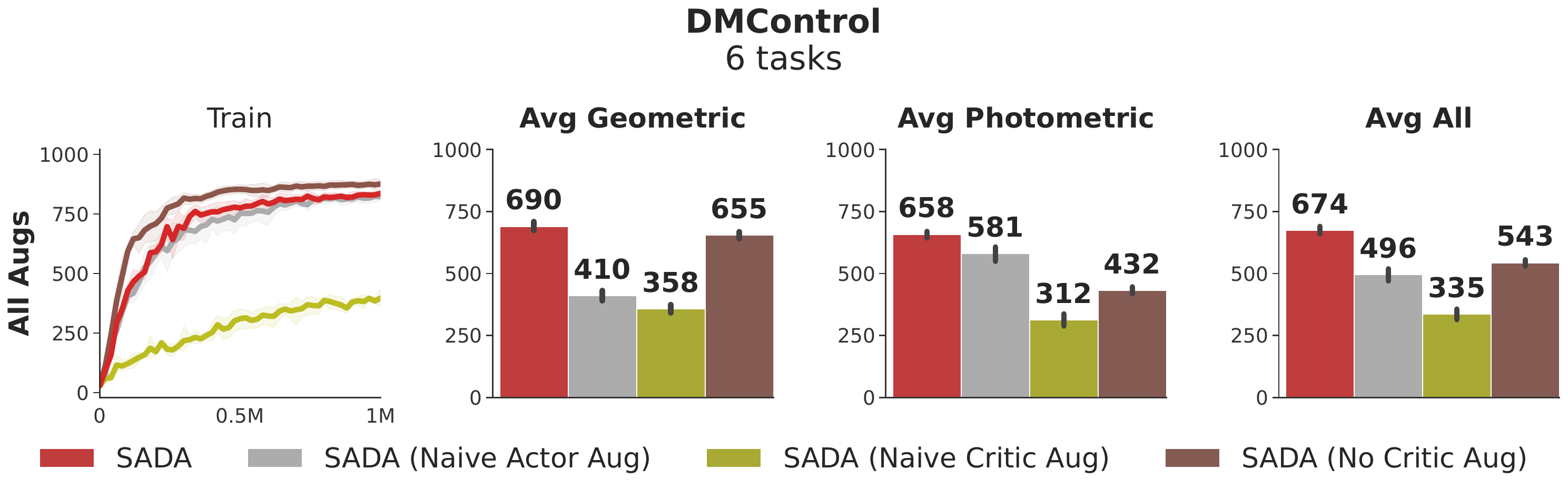}
    \vspace{-0.2in}
    \caption{\textbf{Ablations.} Episode reward on DMC-GB2 when trained under all augmentations, averaged across all DMControl tasks. SADA (Naive Actor Aug) and SADA (Naive Critic Aug) correspond to naive application of augmentation to the actor and the critic respectively. SADA (No Critic Aug) corresponds to applying augmentation only to the actor without any application of augmentation to the critic. More details in Appendix \ref{subsection:app_ablations}. Mean and 95\% CI over 5 random seeds.}
    \label{fig:abl_main}
    \vspace{-0.1in}
\end{figure}
We ablate each of our \emph{design choices}, evaluating methods under all augmentations on all DMC-GB2 test sets, and show results in Figure \ref{fig:abl_main}. Naively applying augmentation to the actor or the critic, as displayed in SADA (Naive Actor Aug) and SADA (Naive Critic Aug) respectively, leads to deteriorated performance. As for SADA (No Critic Aug), we only apply augmentation to the actor using SADA's objective without any application of augmentation to the critic. SADA (No Critic Aug) displays impressive geometric robustness and training sample efficiency, but lacks in photometric robustness. If a user is only interested in geometric robustness, SADA (No Critic Aug) provides commendable geometric robustness. Overall, each of our design choices play a key role in establishing the superiority of SADA in all applications of data augmentation.

\begin{figure}[H]
    \centering
    \begin{minipage}[b]{0.47\textwidth}
        \includegraphics[width=\linewidth,trim={0 0 0 2.7cm},clip]{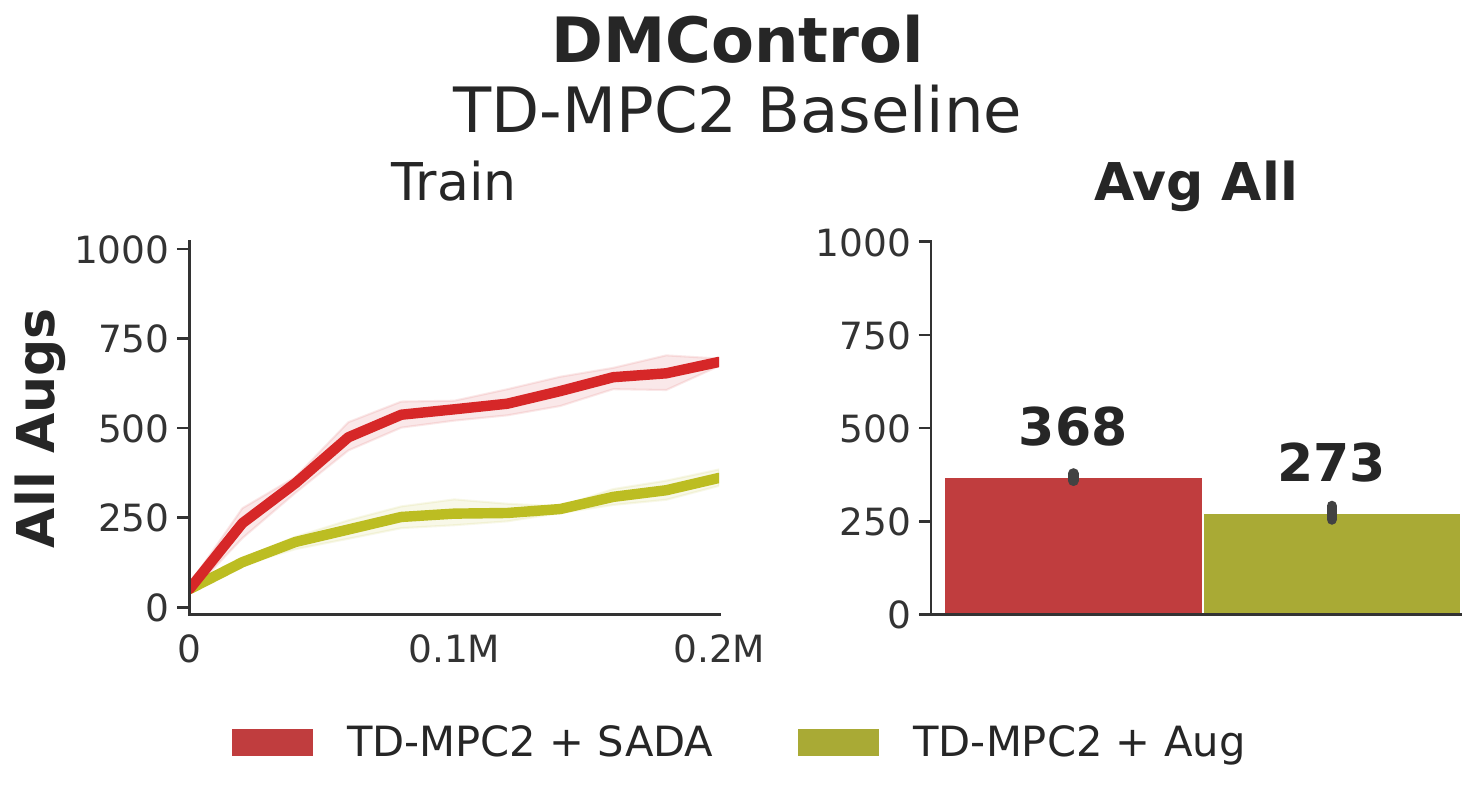}
        \caption{\textbf{TD-MPC2 Baseline.}  Episode reward on DMC-GB2 when trained under all augmentations with a TD-MPC2 backbone, averaged across all DMControl tasks. Mean and 95\% CI over 5 seeds.}
        \label{fig:tdmpc_main}
    \end{minipage}
    \hfill
    \begin{minipage}[b]{0.47\textwidth}    
        \includegraphics[width=\linewidth,trim={0 0 0 2.5cm},clip]{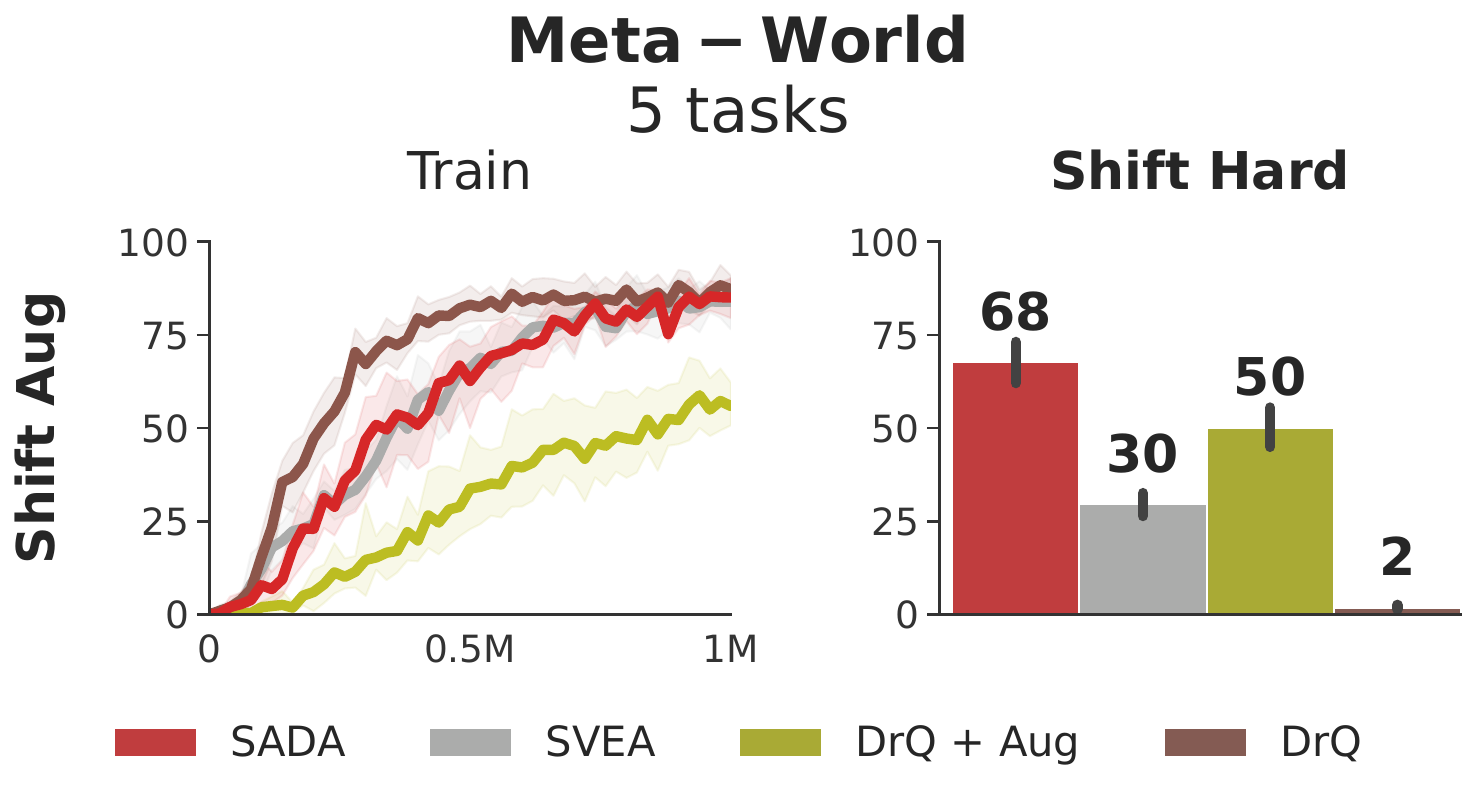}
        \caption{\textbf{Meta-World.} Success rate (\%) on Shift Hard (Meta-World) distribution when trained under \emph{strong} shift augmentation only, averaged across all Meta-World tasks. Mean and 95\% CI for 5 random seeds.}
        \label{fig:mw_main}
    \end{minipage}

\end{figure}

\textbf{Generality.} To demonstrate the generality of our approach, we swap our DrQ backbone with TD-MPC2 \citep{Hansen2022tdmpc, hansen2024tdmpc2}, a state-of-the-art model-based RL algorithm; results are shown in Figure \ref{fig:tdmpc_main}. We observe that SADA similarly improves training stability and generalization of TD-MPC2 on DMC-GB2.

We further evaluate our DrQ-based SADA on our Meta-World setup (see Appendix \ref{subsection:app_mw}), and showcase the results in Figure \ref{fig:mw_main}. Even on Meta-World, SADA surpasses all other baselines in terms of success rate, all while achieving similar training sample efficiency to its unaugmented DrQ baseline. This asserts that SADA can be \emph{readily applied} to diverse tasks, environments, and backbones, and can be used a generic data augmentation strategy for modern visual based reinforcement learning.

\vspace{-0.1in}
\section{Summary}
Throughout this work, we give an overview of data augmentation within visual RL, highlighting the shortcomings of previous work, its implications, and presenting SADA, a generic data augmentation recipe for modern visual based reinforcement learning. We empirically prove SADA's superiority to previous methods and provide a deep analysis of its effectiveness. Concurrently, we curated a comprehensive visual generalization benchmark, DMC-GB2, which we make publicly available at \href{https://aalmuzairee.github.io/SADA}{https://aalmuzairee.github.io/SADA}, with the aim of furthering research efforts within visual RL.


\bibliography{main}
\bibliographystyle{rlc}

\clearpage
\newpage

\appendix

\section{Setup and Implementation}
\label{sec:app_back}
\subsection{Hyper-parameters}
\label{subsection:app_hparams}

\begin{table}[H]
\centering
\begin{tabular}{lc}
\toprule
Parameter        & Setting \\
\midrule
Replay buffer capacity & $10^6$ \\
Action repeat & $2$ \\
Frame stack & $3$ \\
Seed frames & $4000$ \\
Exploration steps & $2000$ \\
Mini-batch size & $256$ \\
Discount $\gamma$ & $0.99$ \\
Optimizer & Adam \\
Learning rate & $5\times10^{-4}$ \\
Agent update frequency & $2$ \\
Critic Q-function soft-update rate $\tau$ & $0.01$ \\
Features dim. & $50$ \\
Hidden dim. & $1024$ \\
Actor log stddev bounds & $[-10,2]$ \\
Init temperature & $0.1$ \\
\multirow{3}{*}{Strong Augmentations}& Max Random Shift Pixels: $16\times16$\\ & Max Random Rotation Degrees: $180 ^{\circ}$\\& Random Overlay Alpha: 0.5 \\
\bottomrule
\end{tabular}
\caption{\label{table:hparams} The default set of hyper-parameters used in our experiments.}

\end{table}

\subsection{Training and Evaluation Setup}
\label{subsection:app_train_setup}
\textbf{DMControl.} Each episode length is set to 1000 environment steps. We train all models for 1M environment steps, evaluating on the training environment every 20,000 environment steps for 10 episodes. Post training, we evaluate trained agents on each level of our test suite for 100 episodes and
report our mean episode reward. We consider six tasks defined below:\\
\begin{figure}[htbp]
    \centering
    \captionsetup{font=small, skip=2pt} 
    \caption*{\emph{Table 2.} \textbf{DMControl}. Task observations are rgb frames of dimensionality $\mathbb{R}^{\mathbf{(3\times84\times84)}} $. We use frame stacking of the three most recent rgb frames such that the observation dimensionality becomes $\mathbb{R}^{(\mathbf{9\times84\times84)}} $. Task difficulty is based on the difficulty classification defined in \citet{yarats2021reinforcement}.}
    \vspace{0.2cm}
    \begin{minipage}[b]{0.4\linewidth}
        \centering
        \resizebox{\linewidth}{!}{%
        \begin{tabular}{ c c c c}
            \toprule
            \textbf{Tasks} & \textbf{Action Dim} & \textbf{Difficulty}\\
            \toprule
            Walker Walk & 6 & Easy \\
            \midrule
            Walker Stand & 6 & Easy \\
            \midrule
            Cheetah Run & 6 & Medium \\
            \midrule
            Finger Spin & 2 & Easy \\
            \midrule
            Cartpole Swingup & 1 & Easy \\
            \midrule
            Cup Catch & 2 & Easy \\
            \bottomrule
        \end{tabular}
        }
    \end{minipage}
\end{figure}

\textbf{Meta-World.} Each episode length is set to 200 environment steps. We train all models for 1M environment steps. Every 20,000 environment steps, we evaluate for 50 episodes and report the mean success rate. At the end of training we evaluate on the test environments for 50 episodes as well, and report the mean success rate. We use the same camera setup as \citet{seo2023masked} and consider five tasks defined below:
\begin{figure}[htbp]
    \centering
    \captionsetup{font=small, skip=2pt} 
    \caption*{\emph{Table 3.} \textbf{Meta-World}. Task observations are rgb frames of dimensionality $\mathbb{R}^{\mathbf{(3\times84\times84)}} $. We use frame stacking of the three most recent rgb frames such that the observation dimensionality becomes $\mathbb{R}^{(\mathbf{9\times84\times84)}} $. Task difficulty is based on the difficulty classification defined in \citet{seo2023masked}.}
    \vspace{0.2cm}
    \begin{minipage}[b]{0.4\linewidth}
        \centering
        \resizebox{\linewidth}{!}{%
        \begin{tabular}{ c c c c}
            \toprule
            \textbf{Tasks} & \textbf{Action Dim} & \textbf{Difficulty}\\
            \toprule
            Door Open & 4 & Easy \\
            \midrule
            Peg Unplug Side & 4 & Easy \\
            \midrule
            Sweep Into & 4 & Medium \\
            \midrule
            Basketball & 4 & Medium \\
            \midrule
            Push & 4 & Hard \\
            \bottomrule
        \end{tabular}
        }
    \end{minipage}
\end{figure}

\subsection{SAC Based Formulation}
\label{subsection:app_sac}

In the following section, we formulate our objective in the context of our base algorithm, Soft Actor Critic \citep{haarnoja2018soft}, but we stress that these changes are applicable to \emph{any} actor critic framework.
The actor update objective for SAC with a learned temperature $\alpha$ thus becomes:
\begin{align}
    \mathcal{L}_{\pi_\phi}^\textbf{SADA}(\mathcal{D}) = \mathbb{E}_{\mathbf{o}_t\sim \mathcal{D}}[D_{KL}(\pi_\phi(\cdot|\mathbf{p}_t))||\exp\{\frac{1}{\alpha}Q_\theta(\mathbf{m}_t,\cdot)\})]. \\
    \mathcal{L}_\alpha^\textbf{SADA}(\mathcal{D}) = \mathbb{E}_{\substack{\mathbf{o}_t \sim \mathcal{D} \\ \mathbf{a_t} \sim \pi_\phi(\cdot|\mathbf{p}_t)}}[-\alpha \log \pi_\phi(\mathbf{a_t}|\mathbf{p}_t) - \alpha \bar{\mathcal{H}}],    
\end{align}
where $\mathbf{p}_t = f_\xi(\left[ \mathbf{o}_t,\mathbf{o}_t^{\text{aug}}\right]_\text{N})$,~ $\mathbf{m}_t = f_\xi(\left[ \mathbf{o}_t,\mathbf{o}_t\right]_\text{N})$, ~ and $\mathbf{o}_t^{\text{aug}} = \text{aug}(\mathbf{o}_t,v_t), v_t \sim V$. 
We use $f_\xi$ to denote the CNN encoder, and $[\cdot]_\text{N}$ to denote concatenation for batch size of dimensionality N where $\mathbf{o}_t, \mathbf{o}_t^\text{aug} \in \mathbb{R}^{\text{N}\times \text{C}\times\text{H}\times\text{W}}$. We use aug() as the augmentation operator where we stochastically sample from the augmentation distribution $\mathcal{V}$ and apply it to the input. 

On the critic's side, the critic's target prediction is unaltered:
\begin{align}
  q_t^{tgt} = r(\mathbf{o}_t, \mathbf{a_t}) + \gamma \text{max}_{\mathbf{a'_t}}Q_{\overline{\theta}}(f_\xi(\mathbf{o}_{t+1}),\mathbf{a'})
\end{align}
while the critic's objective is changed to become:
\begin{align}
    \mathcal{L}_{Q_\theta}^\textbf{SADA}(\mathcal{D}) &= \mathbb{E}_{\mathbf{o}_t,\mathbf{a_t}, \mathbf{r_t}, \mathbf{o}_{t+1}\sim\mathcal{D}} \Bigr[\big\|Q_\theta(\mathbf{p}_t,\mathbf{a_t}) - \mathbf{y}_t\big\|_2\Bigl] 
\end{align}
where $\mathbf{p}_t = f_\xi(\left[ \mathbf{o}_t,\mathbf{o}_t^{\text{aug}}\right]_{\text{N}})$,~ and $\mathbf{y}_t = \left[ q_t^{tgt},q_t^{tgt}\right]_{\text{N}}$.

\subsection{Pseudocode}
\label{subsection:app_pseudo}
\begin{algorithm}[H]
\caption{~~Generic \textbf{SADA} Visual Actor Critic Algorithm\\~~({\color{BrickRed}$\blacktriangleright$~naïve augmentation},~~{\color{citecolor}$\blacktriangleright$~our modifications})}
\label{alg:pseudo-code}
\begin{algorithmic}[1]
\algnotext{EndFor}
\Statex $f_\xi$, $\pi_\phi$, $Q_{\theta}$: encoder, policy, and Q-function respectively
\Statex $T$, $\eta$, $\mathcal{D}$, $\tau$: training steps, learning rate, data replay buffer, target update rate
\Statex $\mathrm{aug}, \mathcal{V}$: choice of strong image augmentation, augmentation distribution

\For{each timestep $t=1..T$}
\State  $\mathbf{a_t} \sim \pi(\cdot | \mathbf{o}_{t})$
\State  $\mathbf{o}_{t+1} \sim p(\cdot | \mathbf{o}_{t}, \mathbf{a_t})$
\State $\mathcal{D} \leftarrow \mathcal{D} \cup (\mathbf{o}_t, \mathbf{a_t}, r(\mathbf{o}_t, \mathbf{a_t}), \mathbf{o}_{t+1})$
\State \textsc{UpdateCritic}($\mathcal{D}$) $\hfill$ 
\State \textsc{UpdateActor}($\mathcal{D}$)
\EndFor

\Procedure{UpdateCritic}{$\mathcal{D}$}
    \State $\{\mathbf{o}_{i}, \mathbf{a}_{i}, r(\mathbf{o}_{i}, \mathbf{a}_{i}), \mathbf{o}_{i+1}~|~i = 1...N \} \sim \mathcal{D}$ \hfill $\vartriangleright$~~Sample batch of transitions
    \State {\color{BrickRed}$\mathbf{o}_{i}, ~\mathbf{o}_{i+1} = \mathrm{aug}(\mathbf{o}_{i}, \nu_{i}), ~\mathrm{aug}(\mathbf{o}_{i+1}, \nu_{i'}) ~\nu_{i},\nu_{i'} \sim \mathcal{V}$}
    \State $q_i^{tgt} = r(\mathbf{o}_i, \mathbf{a_i}) + \gamma \text{max}_{\mathbf{a'_i}}Q_{\overline{\theta}}(f_\xi(\mathbf{o}_{i+1}),\mathbf{a'})$ \hfill $\vartriangleright$~~ Compute Q-target
    \State \textcolor{citecolor}{$\mathbf{o}^{\textnormal{aug}}_{i} = \mathrm{aug}(\mathbf{o}_{i}, \nu_{i}),~\nu_{i} \sim \mathcal{V}$} \hfill {\color{citecolor}$\blacktriangleright$~~Apply stochastic data augmentation}
    \State \textcolor{citecolor}{$\mathbf{p}_{i} = \left[ \mathbf{o}_{i}, \mathbf{o}^{\textnormal{aug}}_{i}\right]_{\textnormal{N}}, ~\mathbf{y}_{i} = \left[ q_i^{tgt}, q_i^{tgt}\right]_{\textnormal{N}} $} \hfill {\color{citecolor}$\blacktriangleright$~~Pack data streams}
    \State \textcolor{citecolor}{$\theta \longleftarrow \theta - \eta \nabla_{\theta}  \mathcal{L}^{\textbf{SADA}}_{Q_{\theta}} \left(\mathbf{p}_i,\mathbf{y}_i \right)$} \hfill {\color{citecolor}$\blacktriangleright$~~Update Q-function and encoder}
    \State {$\overline{\theta} \longleftarrow (1-\tau)\overline{\theta} + \tau\theta $} \hfill $\vartriangleright$~~ Update target Q-function weights

\EndProcedure
\Procedure{UpdateActor}{$\mathcal{D}$}
    \State $\{\mathbf{o}_{i}, \mathbf{a}_{i}, r(\mathbf{o}_{i}, \mathbf{a}_{i}), \mathbf{o}_{i+1}~|~i = 1...N \} \sim \mathcal{D}$ \hfill $\vartriangleright$~~Sample batch of transitions
    \State {\color{BrickRed}$\mathbf{o}_{i} = \mathrm{aug}(\mathbf{o}_{i}, \nu_{i}),~\nu_{i} \sim \mathcal{V}$}
    \State \textcolor{citecolor}{$\mathbf{o}^{\textnormal{aug}}_{i} = \mathrm{aug}(\mathbf{o}_{i}, \nu_{i}),~\nu_{i} \sim \mathcal{V}$} \hfill {\color{citecolor}$\blacktriangleright$~~Apply stochastic data augmentation}
    \State \textcolor{citecolor}{$\mathbf{p}_{i} = \left[ \mathbf{o}_{i}, \mathbf{o}^{\textnormal{aug}}_{i}\right]_{\textnormal{N}}, ~\mathbf{m}_{i} = \left[ \mathbf{o}_{i}, \mathbf{o}_{i}\right]_{\textnormal{N}} $} \hfill {\color{citecolor}$\blacktriangleright$~~Pack data streams}
    \State \textcolor{citecolor}{$\phi \longleftarrow \phi - \eta \nabla_{\phi}  \mathcal{L}^{\textbf{SADA}}_{\pi_{\phi}} \left(\mathbf{p}_i,\mathbf{m}_i \right)$} \hfill {\color{citecolor}$\blacktriangleright$~~Update policy}
\EndProcedure
\end{algorithmic}
\end{algorithm}

\clearpage
\section{Extended Analysis}
\label{sec:app_analysis}
\subsection{Ablations}
\label{subsection:app_ablations}

We ablate all our design choices and show the specific modifications in Figure \ref{table:app_ablations}. We refer to SADA's application of augmentation as selective, where not all inputs are augmented. We use 'naive' to refer to a naive application of augmentation, where all inputs are augmented. We use - to denote no application of augmentation. 
\begin{figure}[H]
    \centering
    \begin{minipage}[b]{1.0\linewidth}
        \centering
        \captionsetup{font=small, skip=2pt} 
        \resizebox{\linewidth}{!}{%
        \begin{tabular}{ c c c c c c }
            
            \toprule
            \textbf{Method} & \textbf{Actor Aug} & \textbf{Critic Aug} & \textbf{Avg Geometric} & \textbf{Avg Photometric} & \textbf{Avg All} \\
            \toprule
            SADA & Selective & Selective & \textbf{690} & \textbf{658} & \textbf{674} \\
            \midrule
            SADA (Naive Actor Aug) & Naive & Selective & 410 & 581 & 496 \\
            \midrule
            SADA (Naive Critic Aug) & Selective & Naive & 358 & 312 & 335 \\
            \midrule
            SADA (No Critic Aug) & Selective & - & 655 & 432 & 543 \\
            \midrule
            SVEA & - & Selective & 232 & 527 & 380 \\
            \midrule
            DrQ + Aug & Naive & Naive & 217 & 246 & 231 \\
            \midrule
            DrQ & - & - & 184 & 322 & 253 \\
            \bottomrule
           
        \end{tabular}
        }
    \end{minipage}
\caption{\textbf{Ablations.} Episode reward on DMC-GB2. Methods trained under all augmentations and averaged across all DMControl tasks. Mean and 95\% CI for 5 random seeds.}
\label{table:app_ablations}
\end{figure}

\subsection{Distracting Control Suite Results}
\label{subsection:app_dcs_results}

We train all methods in the DMControl training environments under all strong augmentations, and evaluate them in a zero-shot manner on the Distracting Control Suite. The results are shown below in Figure \ref{fig:app_dcs}, where SADA outperforms all baselines using the current augmentations.

\begin{figure}[H]
    \centering
    \includegraphics[width=0.5\linewidth,trim={0 0 0 4.5cm},clip]{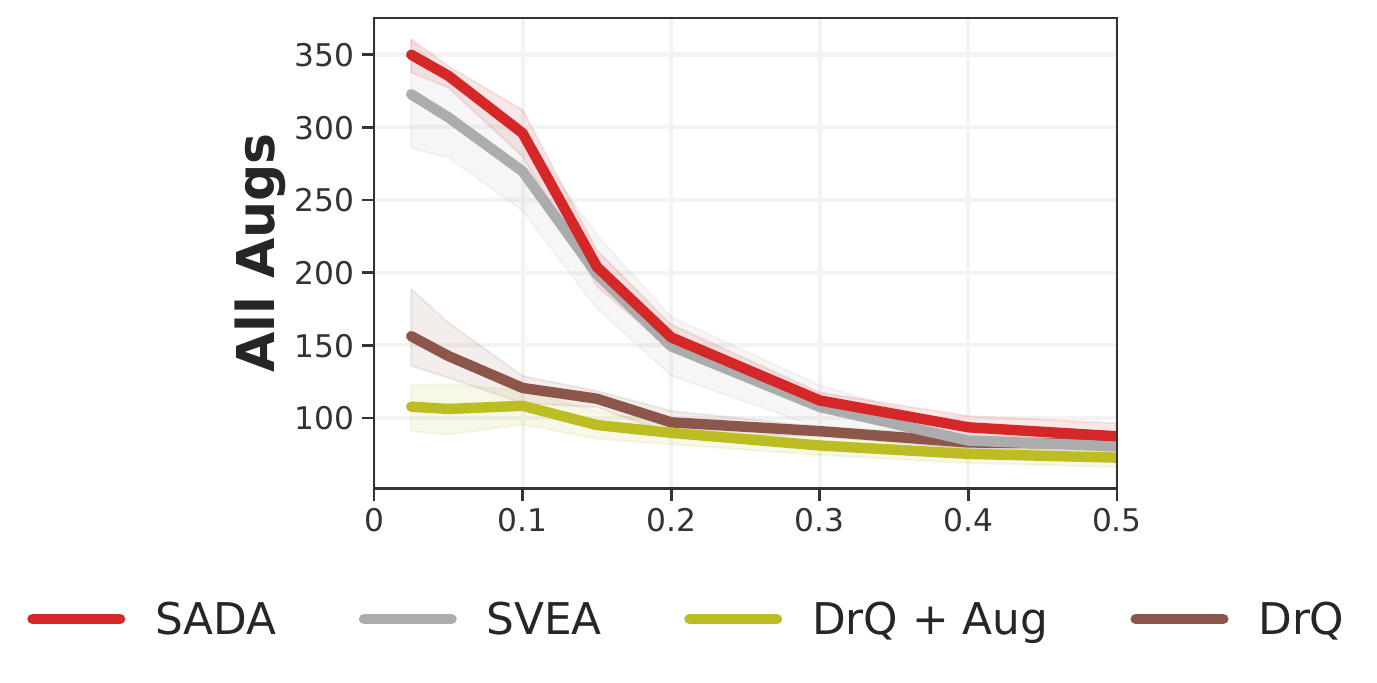}
    \caption{\textbf{Distracting Control Suite.} Episode reward on Distracting Control Suite. Methods trained under all augmentations and averaged across all DMControl tasks. Mean and 95\% CI for 5 random seeds.}
    \label{fig:app_dcs}
\end{figure}

\clearpage
\subsection{T-SNE Visualization}
\label{subsection:app_tsne}

\begin{figure}[H]
    \centering
    \includegraphics[width=0.5\linewidth,trim={0 0 0 2.5cm},clip]{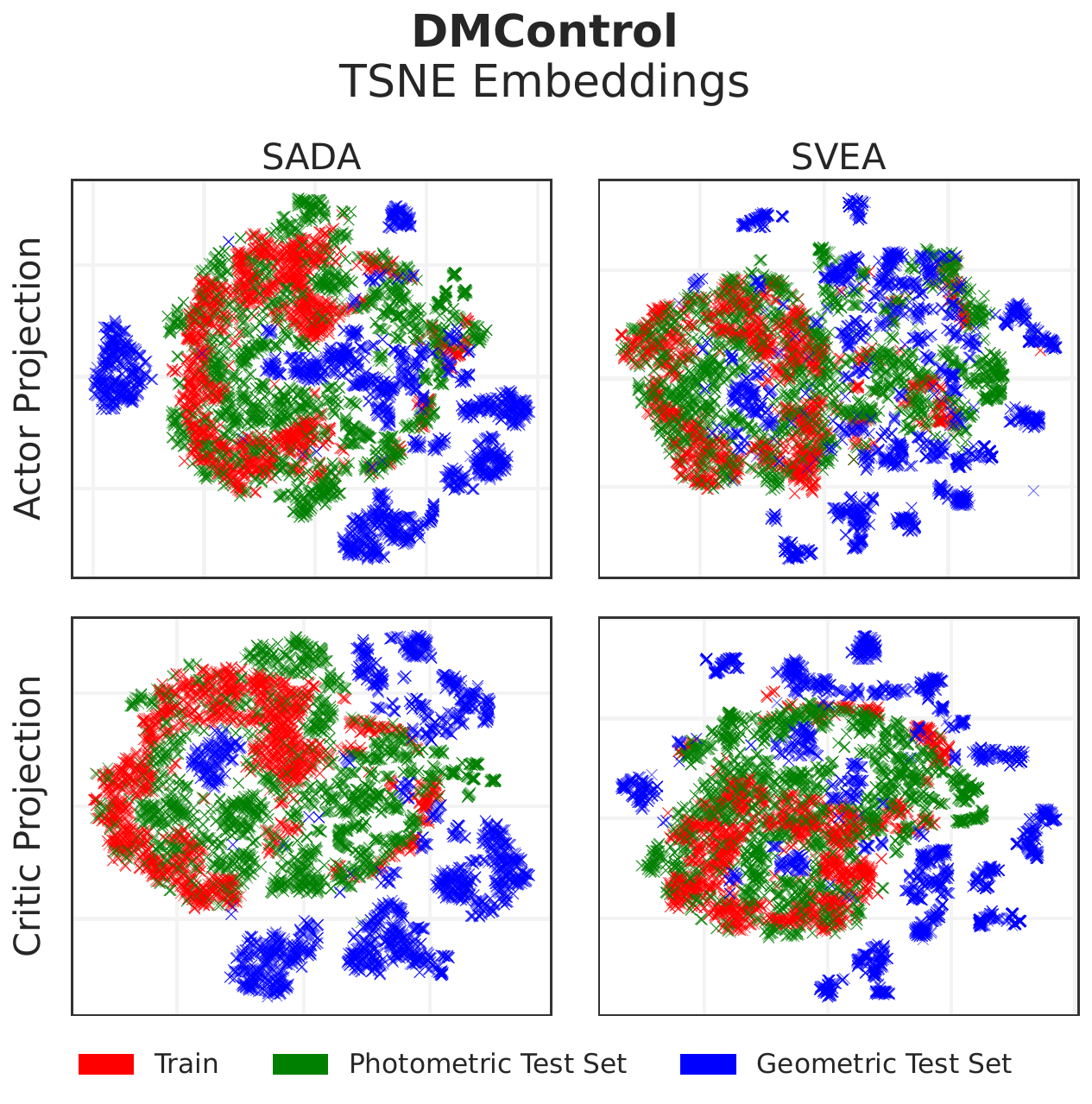}
    \caption{\textbf{T-SNE Embeddings.} We use T-SNE to visualize the projections of converged SADA and SVEA agents trained under all augmentations in the Walker Walk task. }
    \label{fig:app_tsne}
\end{figure}

We visualize the T-SNE projection of converged SADA and SVEA agents in Figure \ref{fig:app_tsne}. Analyzing the graph, we notice a general trend where photometric distributions largely overlap with the training distribution, while geometric distributions seem distant and have little overlap with the training distribution. This asserts the fact presented in Figure \ref{fig:main_fig}, that the CNN encoder can align the photometric augmentations with the training distribution, such that their latent space is similar. On the other hand, geometric augmentations induce changes in the encoder's output embedding that force it to be placed in seperate latent space.

\clearpage
\section{Visuals}
\label{section:app_visuals}

\subsection{Augmentations}
\label{subsection:app_augs}

\begin{figure}[H]
    \centering
    \begin{subfigure}[b]{0.48\textwidth}
        \centering
        \includegraphics[width=\textwidth]{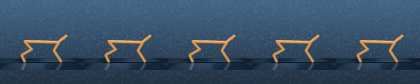}
        \caption{No augmentation (cheetah)}
        \vspace{0.1in}
    \end{subfigure}
    \begin{subfigure}[b]{0.48\textwidth}
        \centering
        \includegraphics[width=\textwidth]{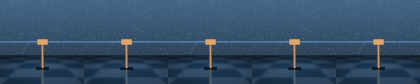}
        \caption{No augmentation (cartpole)}
        \vspace{0.1in}
    \end{subfigure}
    \begin{subfigure}[b]{0.48\textwidth}
        \centering
        \includegraphics[width=\textwidth]{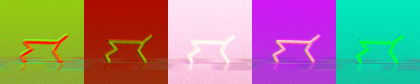}
        \caption{Random convolution (cheetah)}
        \vspace{0.1in}
    \end{subfigure}
    \begin{subfigure}[b]{0.48\textwidth}
        \centering
        \includegraphics[width=\textwidth]{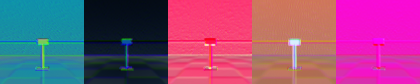}
        \caption{Random convolution (cartpole)}
        \vspace{0.1in}
    \end{subfigure}
    \begin{subfigure}[b]{0.48\textwidth}
        \centering
        \includegraphics[width=\textwidth]{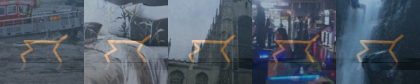}
        \caption{Random overlay (cheetah)}
        \vspace{0.1in}
    \end{subfigure}
    \begin{subfigure}[b]{0.48\textwidth}
        \centering
        \includegraphics[width=\textwidth]{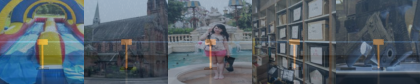}
        \caption{Random overlay (cartpole)}
        \vspace{0.1in}
    \end{subfigure}
    \begin{subfigure}[b]{0.48\textwidth}
        \centering
        \includegraphics[width=\textwidth]{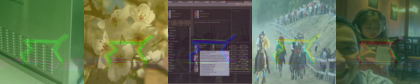}
        \caption{Random convolution and overlay (cheetah)}
        \vspace{0.1in}
    \end{subfigure}
    \begin{subfigure}[b]{0.48\textwidth}
        \centering
        \includegraphics[width=\textwidth]{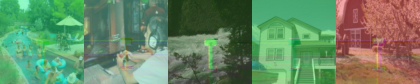}
        \caption{Random convolution and overlay (cartpole)}
        \vspace{0.1in}
    \end{subfigure}
    \begin{subfigure}[b]{0.48\textwidth}
        \centering
        \includegraphics[width=\textwidth]{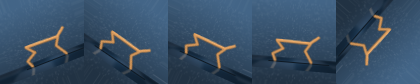}
        \caption{Random rotate (cheetah)}
        \vspace{0.1in}
    \end{subfigure}
    \begin{subfigure}[b]{0.48\textwidth}
        \centering
        \includegraphics[width=\textwidth]{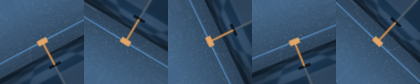}
        \caption{Random rotate (cartpole)}
        \vspace{0.1in}
    \end{subfigure}
    \begin{subfigure}[b]{0.48\textwidth}
        \centering
        \includegraphics[width=\textwidth]{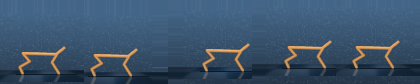}
        \caption{Random shift (cheetah)}
        \vspace{0.1in}
    \end{subfigure}
    \begin{subfigure}[b]{0.48\textwidth}
        \centering
        \includegraphics[width=\textwidth]{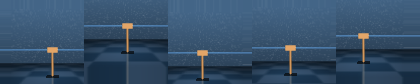}
        \caption{Random shift (cartpole)}
        \vspace{0.1in}
    \end{subfigure}
    \begin{subfigure}[b]{0.48\textwidth}
        \centering
        \includegraphics[width=\textwidth]{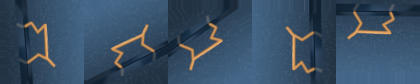}
        \caption{Random rotate and shift (cheetah)}
        \vspace{0.1in}
    \end{subfigure}
    \begin{subfigure}[b]{0.48\textwidth}
        \centering
        \includegraphics[width=\textwidth]{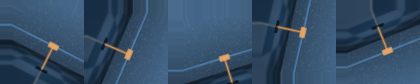}
        \caption{Random rotate and shift (cartpole)}
        \vspace{0.1in}
    \end{subfigure}
    \caption{\textbf{Data augmentation}. Visualizations of data augmentations applied in this study. Left column contains samples from the \textit{Cheetah Run} task, and right column contains samples from the \textit{Cartpole Swingup} task. Sets (c)-(h) constitute of photometric augmentations while sets (i)-(n) constitute of geometric augmentations.}
    \label{fig:data-aug-visualization}
\end{figure}

\subsection{DeepMind Control Suite}
\label{subsection:app_dmc}

\begin{figure}[H]
    \centering
    \begin{subfigure}[b]{0.48\textwidth}
        \centering
        \includegraphics[width=\textwidth]{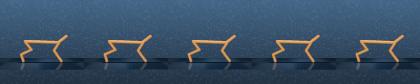}
        \caption{Training environment (cheetah)}
        \vspace{0.1in}
    \end{subfigure}
    \begin{subfigure}[b]{0.48\textwidth}
        \centering
        \includegraphics[width=\textwidth]{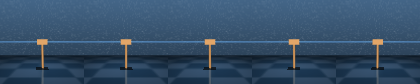}
        \caption{Training environment (cartpole)}
        \vspace{0.1in}
    \end{subfigure}
    \vspace{-0.1in}
    \caption{\textbf{DMControl Train environment.} \textit{(Left)} \textit{Cheetah Run} task. \textit{(Right)} \textit{Cartpole Swingup} task. }
    \label{fig:train-dmc-vis}
    \vspace{-0.15in}
\end{figure}

\begin{figure}[H]
    \centering
    \begin{subfigure}[b]{0.48\textwidth}
        \centering
        \includegraphics[width=\textwidth]{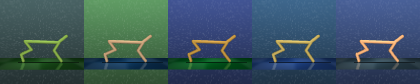}
        \caption{\texttt{color\_easy} environment (cheetah)}
        \vspace{0.1in}
    \end{subfigure}
    \begin{subfigure}[b]{0.48\textwidth}
        \centering
        \includegraphics[width=\textwidth]{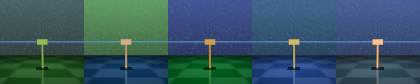}
        \caption{\texttt{color\_easy} environment (cartpole)}
        \vspace{0.1in}
    \end{subfigure}
    \begin{subfigure}[b]{0.48\textwidth}
        \centering
        \includegraphics[width=\textwidth]{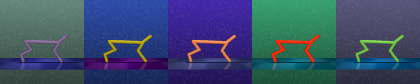}
        \caption{\texttt{color\_hard} environment (cheetah)}
        \vspace{0.1in}
    \end{subfigure}
    \begin{subfigure}[b]{0.48\textwidth}
        \centering
        \includegraphics[width=\textwidth]{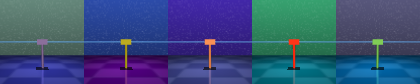}
        \caption{\texttt{color\_hard} environment (cartpole)}
        \vspace{0.1in}
    \end{subfigure}
    \begin{subfigure}[b]{0.48\textwidth}
        \centering
        \includegraphics[width=\textwidth]{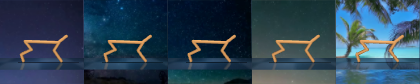}
        \caption{\texttt{video\_easy} environment (cheetah)}
        \vspace{0.1in}
    \end{subfigure}
    \begin{subfigure}[b]{0.48\textwidth}
        \centering
        \includegraphics[width=\textwidth]{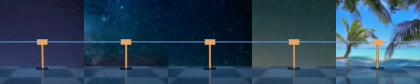}
        \caption{\texttt{video\_easy} environment (cartpole)}
        \vspace{0.1in}
    \end{subfigure}
    \begin{subfigure}[b]{0.48\textwidth}
        \centering
        \includegraphics[width=\textwidth]{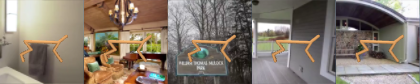}
        \caption{\texttt{video\_hard} environment (cheetah)}
        \vspace{0.1in}
    \end{subfigure}
    \begin{subfigure}[b]{0.48\textwidth}
        \centering
        \includegraphics[width=\textwidth]{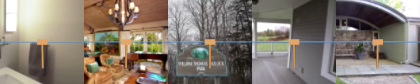}
        \caption{\texttt{video\_hard} environment (cartpole)}
        \vspace{0.1in}
    \end{subfigure}
    \begin{subfigure}[b]{0.48\textwidth}
        \centering
        \includegraphics[width=\textwidth]{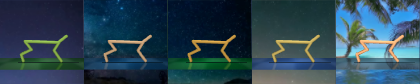}
        \caption{\texttt{color\_video\_easy} environment (cheetah)}
        \vspace{0.1in}
    \end{subfigure}
    \begin{subfigure}[b]{0.48\textwidth}
        \centering
        \includegraphics[width=\textwidth]{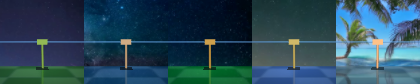}
        \caption{\texttt{color\_video\_easy} environment (cartpole)}
        \vspace{0.1in}
    \end{subfigure}
    \begin{subfigure}[b]{0.48\textwidth}
        \centering
        \includegraphics[width=\textwidth]{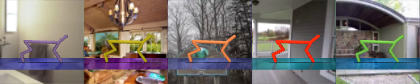}
        \caption{\texttt{color\_video\_hard} environment (cheetah)}
        \vspace{0.1in}
    \end{subfigure}
    \begin{subfigure}[b]{0.48\textwidth}
        \centering
        \includegraphics[width=\textwidth]{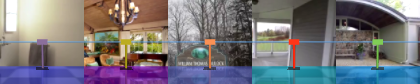}
        \caption{\texttt{color\_video\_hard} environment (cartpole)}
        \vspace{0.1in}
    \end{subfigure}
    \vspace{-0.1in}
    \caption{\textbf{DMC-GB2 Photometric Test Set.} Visualizations from the 6 photometric test distributions in DMC-GB2.}
    \label{fig:photo-dmc-vis}
    \vspace{-0.15in}
\end{figure}

\begin{figure}[H]
    \centering
    \begin{subfigure}[b]{0.48\textwidth}
        \centering
        \includegraphics[width=\textwidth]{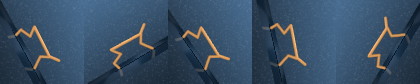}
        \caption{\texttt{rotate\_easy} environment (cheetah)}
        \vspace{0.1in}
    \end{subfigure}
    \begin{subfigure}[b]{0.48\textwidth}
        \centering
        \includegraphics[width=\textwidth]{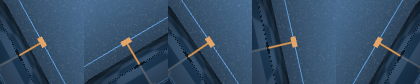}
        \caption{\texttt{rotate\_easy} environment (cartpole)}
        \vspace{0.1in}
    \end{subfigure}
    \begin{subfigure}[b]{0.48\textwidth}
        \centering
        \includegraphics[width=\textwidth]{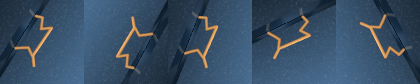}
        \caption{\texttt{rotate\_hard} environment (cheetah)}
        \vspace{0.1in}
    \end{subfigure}
    \begin{subfigure}[b]{0.48\textwidth}
        \centering
        \includegraphics[width=\textwidth]{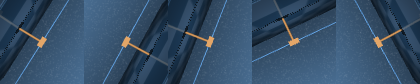}
        \caption{\texttt{rotate\_hard} environment (cartpole)}
        \vspace{0.1in}
    \end{subfigure}
    \begin{subfigure}[b]{0.48\textwidth}
        \centering
        \includegraphics[width=\textwidth]{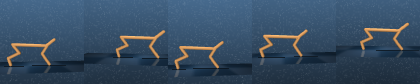}
        \caption{\texttt{shift\_easy} environment (cheetah)}
        \vspace{0.1in}
    \end{subfigure}
    \begin{subfigure}[b]{0.48\textwidth}
        \centering
        \includegraphics[width=\textwidth]{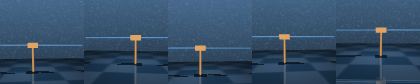}
        \caption{\texttt{shift\_easy} environment (cartpole)}
        \vspace{0.1in}
    \end{subfigure}
    \begin{subfigure}[b]{0.48\textwidth}
        \centering
        \includegraphics[width=\textwidth]{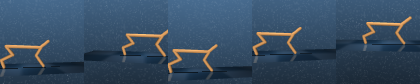}
        \caption{\texttt{shift\_hard} environment (cheetah)}
        \vspace{0.1in}
    \end{subfigure}
    \begin{subfigure}[b]{0.48\textwidth}
        \centering
        \includegraphics[width=\textwidth]{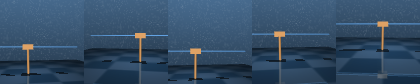}
        \caption{\texttt{shift\_hard} environment (cartpole)}
        \vspace{0.1in}
    \end{subfigure}
    \begin{subfigure}[b]{0.48\textwidth}
        \centering
        \includegraphics[width=\textwidth]{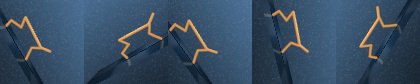}
        \caption{\texttt{rotate\_shift\_easy} environment (cheetah)}
        \vspace{0.1in}
    \end{subfigure}
    \begin{subfigure}[b]{0.48\textwidth}
        \centering
        \includegraphics[width=\textwidth]{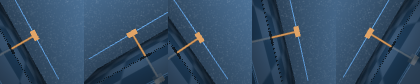}
        \caption{\texttt{rotate\_shift\_easy} environment (cartpole)}
        \vspace{0.1in}
    \end{subfigure}
    \begin{subfigure}[b]{0.48\textwidth}
        \centering
        \includegraphics[width=\textwidth]{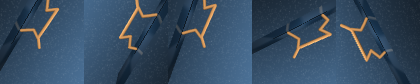}
        \caption{\texttt{rotate\_shift\_hard} environment (cheetah)}
        \vspace{0.1in}
    \end{subfigure}
    \begin{subfigure}[b]{0.48\textwidth}
        \centering
        \includegraphics[width=\textwidth]{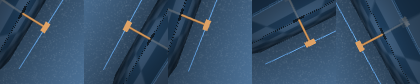}
        \caption{\texttt{rotate\_shift\_easy} environment (cartpole).}
        \vspace{0.1in}
    \end{subfigure}
    \vspace{-0.1in}
    \caption{\textbf{DMC-GB2 Geometric Test Set.} Visualizations from the 6 geometric test distributions in DMC-GB2.}
    \label{fig:geo-dmc-vis}
    \vspace{-0.15in}
\end{figure}

\subsection{Meta-World}
\label{subsection:app_mw}

\begin{figure}[H]
    \centering
    \begin{subfigure}[b]{0.48\textwidth}
        \centering
        \includegraphics[width=\textwidth]{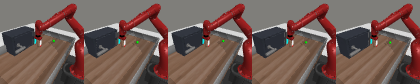}
        \caption{Training environment (door-open)}
        \vspace{0.1in}
    \end{subfigure}
    \begin{subfigure}[b]{0.48\textwidth}
        \centering
        \includegraphics[width=\textwidth]{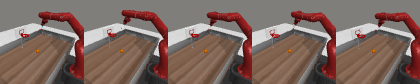}
        \caption{Training environment (basketball)}
        \vspace{0.1in}
    \end{subfigure}
    \vspace{-0.1in}
    \caption{\textbf{Meta-World Train environment.} \textit{(Left)} \textit{Door Open} task. \textit{(Right)} \textit{Basketball} task. }
    \label{fig:train-mw-vis}
    \vspace{-0.15in}
\end{figure}

\begin{figure}[H]
    \centering
    \begin{subfigure}[b]{0.48\textwidth}
        \centering
        \includegraphics[width=\textwidth]{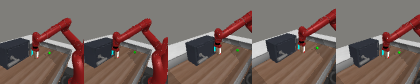}
        \caption{\texttt{shift\_hard} environment (door-open)}
        \vspace{0.1in}
    \end{subfigure}
    \begin{subfigure}[b]{0.48\textwidth}
        \centering
        \includegraphics[width=\textwidth]{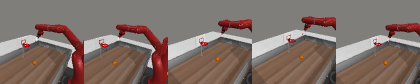}
        \caption{\texttt{shift\_hard} environment (basketball)}
        \vspace{0.1in}
    \end{subfigure}
    \vspace{-0.1in}
    \caption{\textbf{Meta-World Test environment.} Geometric test distribution in Meta-World.}    
    \label{fig:test-mw-vis}
    \vspace{-0.15in}
\end{figure}

\clearpage
\section{Extended Results}
\label{sec:app_extended_results}
\vspace{-0.1cm}
\subsection{DeepMind Control Suite Results}
\label{subsection:app_dmc_extended_results}

\begin{figure}[htbp]
    \vspace{-0.1cm}
    \centering
    \begin{minipage}[b]{0.48\linewidth}
        \centering
        \captionsetup{font=small, skip=2pt} 
        \caption*{a) \textbf{Rotate Easy}}
        \resizebox{\linewidth}{!}{%
            \begin{tabular}{ccccc}
            \toprule
             &  DrQ &  DrQ + Aug & SVEA & SADA \\
            \midrule
                Walker Walk & 232±23 & 166±33 & 278±21 & \textbf{808±90} \textbf{*} \\
                Walker Stand & 408±24 & 329±113 & 505±24 & \textbf{958±6} \textbf{*} \\
                Cheetah Run & 89±10 & 84±44 & 127±26 & \textbf{302±57} \textbf{*} \\
                Finger Spin & 116±39 & 618±80 & 148±15 & \textbf{870±152} \textbf{*} \\
                Cartpole Swingup & 228±29 & 219±17 & 295±23 & \textbf{743±56} \textbf{*} \\
                Cup Catch & 409±45 & 111±40 & 408±150 & \textbf{909±30} \textbf{*} \\
            \bottomrule
            \end{tabular}
        }
    \end{minipage}
    \hspace{0.02\linewidth} 
    \begin{minipage}[b]{0.48\linewidth}
        \centering
        \captionsetup{font=small, skip=2pt} 
        \caption*{b) \textbf{Rotate Hard}}
        \resizebox{\linewidth}{!}{%
            \begin{tabular}{ccccc}
            \toprule
              &  DrQ &  DrQ + Aug & SVEA & SADA \\
            \midrule
                Walker Walk & 133±11 & 147±22 & 154±7 & \textbf{799±89} \textbf{*} \\
                Walker Stand & 268±11 & 288±79 & 330±20 & \textbf{960±9} \textbf{*} \\
                Cheetah Run & 46±3 & 86±48 & 72±16 & \textbf{290±80} \textbf{*} \\
                Finger Spin & 59±20 & 603±116 & 75±7 & \textbf{862±149} \textbf{*}\\
                Cartpole Swingup & 178±15 & 211±19 & 219±9 & \textbf{746±57} \textbf{*}\\
                Cup Catch & 277±38 & 107±46 & 241±76 & \textbf{908±39} \textbf{*}\\
            \bottomrule
            \end{tabular}
        }
    \end{minipage}
    \vspace{4pt}

    \centering
    \begin{minipage}[b]{0.48\linewidth}
        \centering
        \captionsetup{font=small, skip=2pt} 
        \caption*{c) \textbf{Shift Easy}}
        \resizebox{\linewidth}{!}{%
            \begin{tabular}{ccccc}
            \toprule
              &  DrQ &  DrQ + Aug & SVEA & SADA \\
            \midrule
                Walker Walk & 63±8 & 153±32 & 288±36 & \textbf{824±95} \textbf{*}\\
                Walker Stand & 299±73 & 307±89 & 656±53 & \textbf{962±5} \textbf{*}\\
                Cheetah Run & 35±7 & 104±45 & 90±28 & \textbf{348±27} \textbf{*}\\
                Finger Spin & 287±84 & 772±23 & 386±47 & \textbf{903±152} \\
                Cartpole Swingup & 274±43 & 212±20 & 421±80 & \textbf{798±33} \textbf{*}\\
                Cup Catch & 884±77 & 128±60 & 771±353 & \textbf{947±15} \\
            \bottomrule
            \end{tabular}
        }
    \end{minipage}
    \hspace{0.02\linewidth} 
    \begin{minipage}[b]{0.48\linewidth}
        \centering
        \captionsetup{font=small, skip=2pt} 
        \caption*{d) \textbf{Shift Hard}}
        \resizebox{\linewidth}{!}{%
            \begin{tabular}{ccccc}
            \toprule
              &  DrQ &  DrQ + Aug & SVEA & SADA \\
            \midrule
                Walker Walk & 36±2 & 93±25 & 58±8 & \textbf{641±139} \textbf{*}\\
                Walker Stand & 161±12 & 251±59 & 228±30 & \textbf{870±38} \textbf{*}\\
                Cheetah Run & 11±4 & 54±30 & 23±15 & \textbf{284±26} \textbf{*}\\
                Finger Spin & 3±2 & 573±38 & 13±15 & \textbf{802±112} \textbf{*}\\
                Cartpole Swingup & 206±31 & 189±29 & 284±53 & \textbf{719±59} \textbf{*}\\
                Cup Catch & 676±91 & 131±50 & 674±284 & \textbf{871±62} \\
            \bottomrule
            \end{tabular}
        }
    \end{minipage}
    \vspace{4pt}

    \centering
    \begin{minipage}[b]{0.48\linewidth}
        \centering
        \captionsetup{font=small, skip=2pt} 
        \caption*{e) \textbf{Rotate Shift Easy}}
        \resizebox{\linewidth}{!}{%
            \begin{tabular}{ccccc}
            \toprule
              &  DrQ &  DrQ + Aug & SVEA & SADA \\
            \midrule
                Walker Walk & 43±5 & 107±32 & 102±13 & \textbf{663±140} \textbf{*}\\
                Walker Stand & 196±27 & 280±83 & 327±19 & \textbf{897±30} \textbf{*}\\
                Cheetah Run & 12±6 & 50±24 & 25±9 & \textbf{231±44} \textbf{*}\\
                Finger Spin & 2±2 & 381±83 & 3±2 & \textbf{732±93} \textbf{*}\\
                Cartpole Swingup & 139±28 & 189±12 & 195±14 & \textbf{644±71} \textbf{*}\\
                Cup Catch & 353±93 & 131±64 & 369±147 & \textbf{815±70} \textbf{*}\\
            \bottomrule
            \end{tabular}
        }
    \end{minipage}
    \hspace{0.02\linewidth} 
    \begin{minipage}[b]{0.48\linewidth}
        \centering
        \captionsetup{font=small, skip=2pt} 
        \caption*{f) \textbf{Rotate Shift Hard}}
        \resizebox{\linewidth}{!}{%
            \begin{tabular}{ccccc}
            \toprule
              &  DrQ &  DrQ + Aug & SVEA & SADA \\
            \midrule
                Walker Walk & 34±3 & 57±11 & 38±4 & \textbf{307±70} \textbf{*}\\
                Walker Stand & 147±10 & 191±35 & 180±19 & \textbf{652±78} \textbf{*}\\
                Cheetah Run & 6±2 & 19±13 & 13±6 & \textbf{131±18} \textbf{*}\\
                Finger Spin & 1±0 & 155±61 & 0±0 & \textbf{476±46} \textbf{*}\\
                Cartpole Swingup & 111±16 & 172±12 & 149±12 & \textbf{497±33} \textbf{*}\\
                Cup Catch & 189±52 & 144±80 & 204±72 & \textbf{668±102} \textbf{*}\\
            \bottomrule
            \end{tabular}
        }
    \end{minipage}
    \vspace{20pt}

    \centering
    \begin{minipage}[b]{0.48\linewidth}
        \centering
        \captionsetup{font=small, skip=2pt} 
        \caption*{a) \textbf{Color Easy}}
        \resizebox{\linewidth}{!}{%
            \begin{tabular}{ccccc}
            \toprule
             &  DrQ &  DrQ + Aug & SVEA & SADA \\
            \midrule
                Walker Walk & 582±47 & 228±48 & 755±55 & \textbf{837±70} \textbf{*}\\
                Walker Stand & 826±39 & 333±103 & 900±47 & \textbf{965±10} \textbf{*}\\
                Cheetah Run & \textbf{341±42} \textbf{*}& 88±39 & 203±89 & 252±69 \\
                Finger Spin & 795±61 & 693±74 & \textbf{924±33} & 895±162 \\
                Cartpole Swingup & 696±54 & 230±28 & 542±104 & \textbf{704±33} \\
                Cup Catch & 833±37 & 139±62 & 821±322 & \textbf{969±5} \\
            \bottomrule
            \end{tabular}
        }
    \end{minipage}
    \hspace{0.02\linewidth} 
    \begin{minipage}[b]{0.48\linewidth}
        \centering
        \captionsetup{font=small, skip=2pt} 
        \caption*{b) \textbf{Color Hard}}
        \resizebox{\linewidth}{!}{%
            \begin{tabular}{ccccc}
            \toprule
              &  DrQ &  DrQ + Aug & SVEA & SADA \\
            \midrule
                Walker Walk & 265±41 & 238±44 & 667±51 & \textbf{825±72} \textbf{*}\\
                Walker Stand & 527±65 & 355±121 & 861±60 & \textbf{963±7} \textbf{*}\\
                Cheetah Run & 178±25 & 87±35 & 133±73 & \textbf{239±75} \\
                Finger Spin & 466±73 & 661±76 & 802±108 & \textbf{868±150} \\
                Cartpole Swingup & 441±43 & 240±22 & 478±101 & \textbf{716±34} \textbf{*}\\
                Cup Catch & 520±68 & 157±66 & 779±320 & \textbf{961±11} \\
            \bottomrule
            \end{tabular}
        }
    \end{minipage}
    \vspace{4pt}

    \centering
    \begin{minipage}[b]{0.48\linewidth}
        \centering
        \captionsetup{font=small, skip=2pt} 
        \caption*{c) \textbf{Video Easy}}
        \resizebox{\linewidth}{!}{%
            \begin{tabular}{ccccc}
            \toprule
              &  DrQ &  DrQ + Aug & SVEA & SADA \\
            \midrule
                Walker Walk & 390±56 & 132±33 & 788±103 & \textbf{791±56} \\
                Walker Stand & 603±41 & 279±63 & \textbf{945±13} & 923±45 \\
                Cheetah Run & 75±52 & 49±9 & 102±56 & \textbf{121±59} \\
                Finger Spin & 441±39 & 654±88 & 774±137 & \textbf{875±157} \\
                Cartpole Swingup & 375±54 & 204±34 & 427±85 & \textbf{524±49} \textbf{*}\\
                Cup Catch & 523±21 & 150±45 & 736±303 & \textbf{934±23} \\
            \bottomrule
            \end{tabular}
        }
    \end{minipage}
    \hspace{0.02\linewidth} 
    \begin{minipage}[b]{0.48\linewidth}
        \centering
        \captionsetup{font=small, skip=2pt} 
        \caption*{d) \textbf{Video Hard}}
        \resizebox{\linewidth}{!}{%
            \begin{tabular}{ccccc}
            \toprule
              &  DrQ &  DrQ + Aug & SVEA & SADA \\
            \midrule
                Walker Walk & 36±5 & 166±29 & 264±57 & \textbf{270±31} \\
                Walker Stand & 154±17 & 225±47 & 429±95 & \textbf{702±65} \textbf{*}\\
                Cheetah Run & 25±16 & 75±20 & 28±8 & \textbf{82±20} \\
                Finger Spin & 7±4 & 234±29 & 263±123 & \textbf{566±118} \textbf{*}\\
                Cartpole Swingup & 98±21 & 154±26 & 259±32 & \textbf{363±31} \textbf{*}\\
                Cup Catch & 111±31 & 152±55 & 416±252 & \textbf{662±43} \textbf{*}\\
            \bottomrule
            \end{tabular}
        }
    \end{minipage}
    \vspace{4pt}

    \centering
    \begin{minipage}[b]{0.48\linewidth}
        \centering
        \captionsetup{font=small, skip=2pt} 
        \caption*{e) \textbf{Color Video Easy}}
        \resizebox{\linewidth}{!}{%
            \begin{tabular}{ccccc}
            \toprule
              &  DrQ &  DrQ + Aug & SVEA & SADA \\
            \midrule
                Walker Walk & 208±49 & 219±36 & 681±44 & \textbf{791±59} \textbf{*}\\
                Walker Stand & 487±28 & 330±105 & 852±36 & \textbf{945±15} \textbf{*}\\
                Cheetah Run & 60±36 & 64±16 & 100±58 & \textbf{153±64} \\
                Finger Spin & 310±30 & 653±74 & 705±147 & \textbf{850±150} \\
                Cartpole Swingup & 327±43 & 217±23 & 427±86 & \textbf{570±38} \textbf{*}\\
                Cup Catch & 447±61 & 160±54 & 716±318 & \textbf{931±36} \\
            \bottomrule
            \end{tabular}
        }
    \end{minipage}
    \hspace{0.02\linewidth} 
    \begin{minipage}[b]{0.48\linewidth}
        \centering
        \captionsetup{font=small, skip=2pt} 
        \caption*{f) \textbf{Color Video Hard}}
        \resizebox{\linewidth}{!}{%
            \begin{tabular}{ccccc}
            \toprule
              &  DrQ &  DrQ + Aug & SVEA & SADA \\
            \midrule
                Walker Walk & 42±10 & 215±37 & 421±67 & \textbf{686±61} \textbf{*}\\
                Walker Stand & 170±17 & 288±84 & 659±69 & \textbf{906±30} \textbf{*}\\
                Cheetah Run & 26±17 & 82±23 & 44±24 & \textbf{99±43} \\
                Finger Spin & 2±2 & 365±52 & 307±139 & \textbf{633±106} \textbf{*}\\
                Cartpole Swingup & 94±17 & 166±30 & 294±45 & \textbf{426±39} \textbf{*}\\
                Cup Catch & 122±48 & 163±77 & 484±291 & \textbf{697±37} \\
            \bottomrule
            \end{tabular}
        }
    \end{minipage}
    \vspace{10pt}

    \caption{\textbf{DMC-GB2 Overall Robustness Results.} Episode Reward. Methods trained under all (geometric and photometric) augmentations and evaluated on the all DMC-GB2 Test Sets. Mean and Stddev over 5 random seeds. Highest scores in bold.  Asterisk (*) indicates that the method is statistically significantly greater than all compared methods with 95\% confidence.}
    \label{}

\end{figure}

\clearpage

\begin{figure}[H]
    \centering
    \includegraphics[width=1.0\linewidth]{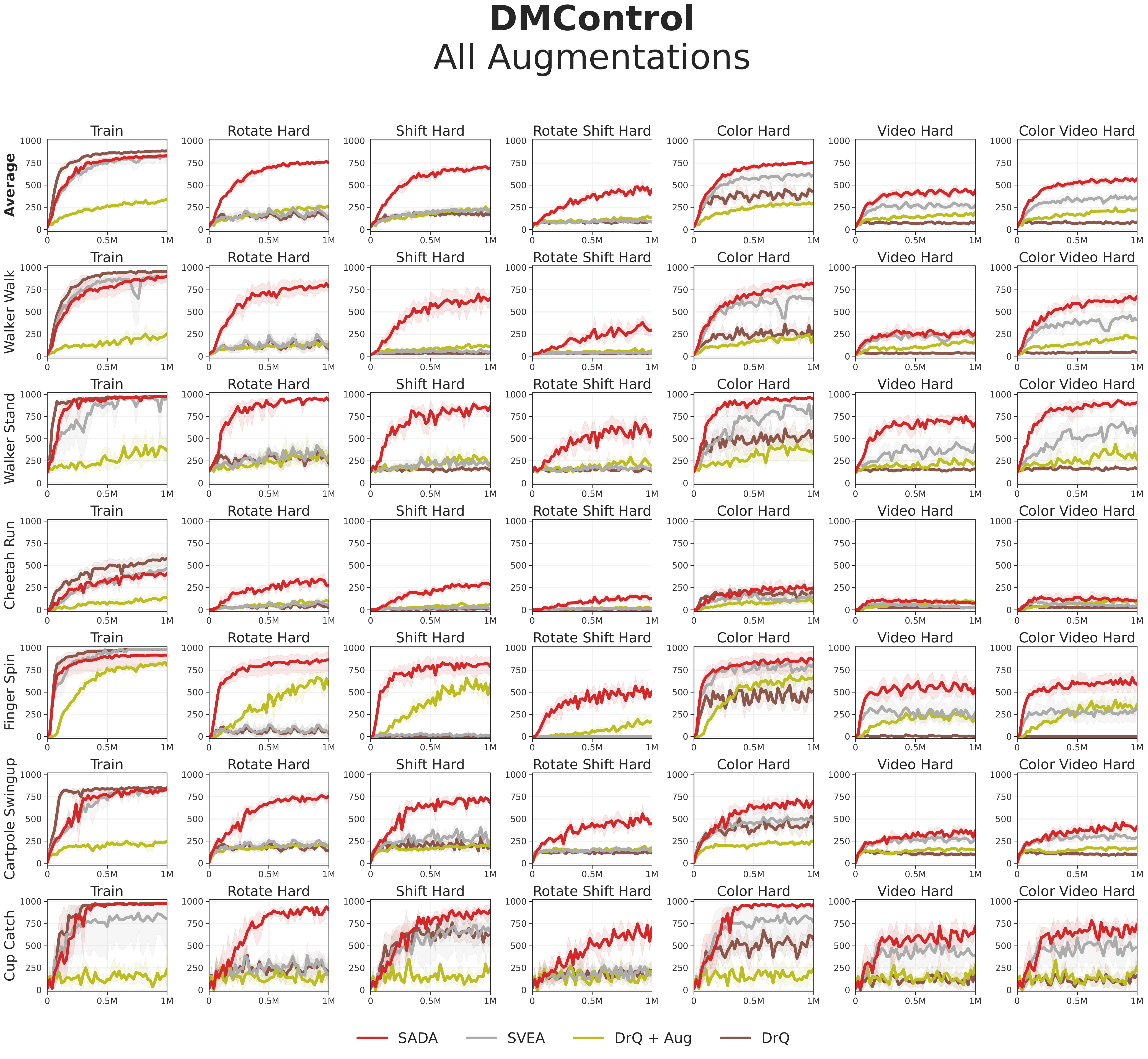}
    \caption{\textbf{DMC-GB2 Overall Robustness Graphs.} Episode Reward. Methods trained under all (geometric and photometric) augmentations and evaluated on all DMC-GB2 Test Sets. Hard levels visualized. Mean and 95\% CI over 5 random seeds.}
    \label{fig:app_dmc_all_graphs}
\end{figure}

\begin{figure}[htbp]
    \vspace{-0.1cm}
    \centering
    \begin{minipage}[b]{0.48\linewidth}
        \centering
        \captionsetup{font=small, skip=2pt} 
        \caption*{a) \textbf{Rotate Easy}}
        \resizebox{\linewidth}{!}{%
            \begin{tabular}{ccccc}
            \toprule
             &  DrQ &  DrQ + Aug & SVEA & SADA \\
            \midrule
            Walker Walk & 232±23 & 431±41 & 426±37 & \textbf{728±369} \\
            Walker Stand & 408±24 & 809±159 & 629±60 & \textbf{968±5} \textbf{*}\\
            Cheetah Run & 89±10 & 180±33 & 147±31 & \textbf{420±76} \textbf{*}\\
            Finger Spin & 116±39 & 829±26 & 257±58 & \textbf{885±150} \\
            Cartpole Swingup & 228±29 & 304±77 & 422±23 & \textbf{801±54} \textbf{*}\\
            Cup Catch & 409±45 & 600±167 & 618±86 & \textbf{803±350} \\
            \bottomrule
            \end{tabular}
        }
    \end{minipage}
    \hspace{0.02\linewidth} 
    \begin{minipage}[b]{0.48\linewidth}
        \centering
        \captionsetup{font=small, skip=2pt} 
        \caption*{b) \textbf{Rotate Hard}}
        \resizebox{\linewidth}{!}{%
            \begin{tabular}{ccccc}
            \toprule
              &  DrQ &  DrQ + Aug & SVEA & SADA \\
            \midrule
            Walker Walk & 133±11 & 416±45 & 228±25 & \textbf{729±367} \\
            Walker Stand & 268±11 & 777±167 & 406±46 & \textbf{961±9} \textbf{*}\\
            Cheetah Run & 46±3 & 168±42 & 86±26 & \textbf{415±76} \textbf{*}\\
            Finger Spin & 59±20 & 820±28 & 128±25 & \textbf{862±158} \\
            Cartpole Swingup & 178±15 & 289±63 & 280±11 & \textbf{798±62} \textbf{*}\\
            Cup Catch & 277±38 & 569±173 & 397±94 & \textbf{797±352} \\
            \bottomrule
            \end{tabular}
        }
    \end{minipage}
    \vspace{4pt}

    \centering
    \begin{minipage}[b]{0.48\linewidth}
        \centering
        \captionsetup{font=small, skip=2pt} 
        \caption*{c) \textbf{Shift Easy}}
        \resizebox{\linewidth}{!}{%
            \begin{tabular}{ccccc}
            \toprule
              &  DrQ &  DrQ + Aug & SVEA & SADA \\
            \midrule
            Walker Walk & 63±8 & 415±61 & 692±67 & \textbf{740±374} \\
            Walker Stand & 299±73 & 822±166 & 765±98 & \textbf{946±15} \\
            Cheetah Run & 35±7 & 179±22 & 133±18 & \textbf{413±72} \textbf{*}\\
            Finger Spin & 287±84 & 678±142 & 460±85 & \textbf{899±136} \textbf{*}\\
            Cartpole Swingup & 274±43 & 288±29 & 564±89 & \textbf{767±57} \textbf{*}\\
            Cup Catch & 884±77 & 695±137 & \textbf{940±43} & 811±343 \\
            \bottomrule
            \end{tabular}
        }
    \end{minipage}
    \hspace{0.02\linewidth} 
    \begin{minipage}[b]{0.48\linewidth}
        \centering
        \captionsetup{font=small, skip=2pt} 
        \caption*{d) \textbf{Shift Hard}}
        \resizebox{\linewidth}{!}{%
            \begin{tabular}{ccccc}
            \toprule
              &  DrQ &  DrQ + Aug & SVEA & SADA \\
            \midrule
            Walker Walk & 36±2 & 304±83 & 154±46 & \textbf{636±324} \textbf{*}\\
            Walker Stand & 161±12 & 671±167 & 387±70 & \textbf{897±31} \textbf{*}\\
            Cheetah Run & 11±4 & 129±14 & 52±18 & \textbf{344±43} \textbf{*}\\
            Finger Spin & 3±2 & 588±204 & 90±48 & \textbf{781±147} \\
            Cartpole Swingup & 206±31 & 216±23 & 318±33 & \textbf{634±104} \textbf{*}\\
            Cup Catch & 676±91 & 604±188 & \textbf{859±77} & 790±348 \\
            \bottomrule
            \end{tabular}
        }
    \end{minipage}
    \vspace{4pt}

    \centering
    \begin{minipage}[b]{0.48\linewidth}
        \centering
        \captionsetup{font=small, skip=2pt} 
        \caption*{e) \textbf{Rotate Shift Easy}}
        \resizebox{\linewidth}{!}{%
            \begin{tabular}{ccccc}
            \toprule
              &  DrQ &  DrQ + Aug & SVEA & SADA \\
            \midrule
            Walker Walk & 43±5 & 316±72 & 228±17 & \textbf{678±345} \textbf{*}\\
            Walker Stand & 196±27 & 705±196 & 489±92 & \textbf{934±22} \textbf{*}\\
            Cheetah Run & 12±6 & 146±14 & 56±16 & \textbf{331±28} \textbf{*}\\
            Finger Spin & 2±2 & 683±169 & 52±39 & \textbf{802±147} \\
            Cartpole Swingup & 139±28 & 257±51 & 269±37 & \textbf{742±58} \textbf{*}\\
            Cup Catch & 353±93 & 586±156 & 589±61 & \textbf{788±347} \\
            \bottomrule
            \end{tabular}
        }
    \end{minipage}
    \hspace{0.02\linewidth} 
    \begin{minipage}[b]{0.48\linewidth}
        \centering
        \captionsetup{font=small, skip=2pt} 
        \caption*{f) \textbf{Rotate Shift Hard}}
        \resizebox{\linewidth}{!}{%
            \begin{tabular}{ccccc}
            \toprule
              &  DrQ &  DrQ + Aug & SVEA & SADA \\
            \midrule
            Walker Walk & 34±3 & 166±75 & 62±12 & \textbf{356±183} \textbf{*}\\
            Walker Stand & 147±10 & 484±189 & 222±33 & \textbf{791±41} \textbf{*}\\
            Cheetah Run & 6±2 & 78±13 & 20±6 & \textbf{180±57} \textbf{*}\\
            Finger Spin & 1±0 & 513±201 & 1±1 & \textbf{663±193} \\
            Cartpole Swingup & 111±16 & 183±28 & 162±14 & \textbf{553±94} \textbf{*}\\
            Cup Catch & 189±52 & 512±159 & 327±49 & \textbf{749±333} \\
            \bottomrule
            \end{tabular}
        }
    \end{minipage}

    \caption{\textbf{DMC-GB2 Geometric Test Set Results.} Episode Reward. Methods trained under geometric augmentations and evaluated on DMC-GB2 Geometric Test Set. Mean and Stddev over 5 random seeds. Highest scores in bold.  Asterisk (*) indicates that the method is statistically significantly greater than all compared methods with 95\% confidence.}
    \label{}

\end{figure}


\begin{figure}[htbp]
    \vspace{-0.1cm}
    \centering
    \begin{minipage}[b]{0.48\linewidth}
        \centering
        \captionsetup{font=small, skip=2pt} 
        \caption*{a) \textbf{Color Easy}}
        \resizebox{\linewidth}{!}{%
            \begin{tabular}{ccccc}
            \toprule
             &  DrQ &  DrQ + Aug & SVEA & SADA \\
            \midrule
            Walker Walk & 582±47 & 911±34 & 841±126 & \textbf{947±26} \\
            Walker Stand & 826±39 & 964±7 & 815±341 & \textbf{975±4} \\
            Cheetah Run & 341±42 & 274±34 & 348±71 & \textbf{368±54} \\
            Finger Spin & 795±61 & 948±51 & 910±142 & \textbf{983±2} \\
            Cartpole Swingup & 696±54 & 626±152 & \textbf{843±16} & 842±19 \\
            Cup Catch & 833±37 & 713±353 & \textbf{976±2} & 973±3 \\
            \bottomrule
            \end{tabular}
        }
    \end{minipage}
    \hspace{0.02\linewidth} 
    \begin{minipage}[b]{0.48\linewidth}
        \centering
        \captionsetup{font=small, skip=2pt} 
        \caption*{b) \textbf{Color Hard}}
        \resizebox{\linewidth}{!}{%
            \begin{tabular}{ccccc}
            \toprule
              &  DrQ &  DrQ + Aug & SVEA & SADA \\
            \midrule
            Walker Walk & 265±41 & 907±31 & 834±127 & \textbf{946±23} \\
            Walker Stand & 527±65 & 963±10 & 813±343 & \textbf{974±2} \\
            Cheetah Run & 178±25 & 273±37 & 333±60 & \textbf{362±48} \\
            Finger Spin & 466±73 & 944±53 & 882±132 & \textbf{980±3} \\
            Cartpole Swingup & 441±43 & 627±149 & 833±15 & \textbf{843±17} \\
            Cup Catch & 520±68 & 722±339 & \textbf{974±2} & 972±4 \\
            \bottomrule
            \end{tabular}
        }
    \end{minipage}
    \vspace{4pt}

    \centering
    \begin{minipage}[b]{0.48\linewidth}
        \centering
        \captionsetup{font=small, skip=2pt} 
        \caption*{c) \textbf{Video Easy}}
        \resizebox{\linewidth}{!}{%
            \begin{tabular}{ccccc}
            \toprule
              &  DrQ &  DrQ + Aug & SVEA & SADA \\
            \midrule
            Walker Walk & 390±56 & 885±41 & 824±143 & \textbf{936±24} \\
            Walker Stand & 603±41 & 964±5 & 813±339 & \textbf{972±2} \\
            Cheetah Run & 75±52 & 264±41 & 298±40 & \textbf{340±50} \\
            Finger Spin & 441±39 & 923±41 & 879±140 & \textbf{972±4} \\
            Cartpole Swingup & 375±54 & 533±157 & \textbf{770±44} & 749±74 \\
            Cup Catch & 523±21 & 690±355 & 947±16 & \textbf{961±5} \\
            \bottomrule
            \end{tabular}
        }
    \end{minipage}
    \hspace{0.02\linewidth} 
    \begin{minipage}[b]{0.48\linewidth}
        \centering
        \captionsetup{font=small, skip=2pt} 
        \caption*{d) \textbf{Video Hard}}
        \resizebox{\linewidth}{!}{%
            \begin{tabular}{ccccc}
            \toprule
              &  DrQ &  DrQ + Aug & SVEA & SADA \\
            \midrule
            Walker Walk & 36±5 & 255±49 & 243±91 & \textbf{329±20} \\
            Walker Stand & 154±17 & 669±79 & 533±203 & \textbf{692±40} \\
            Cheetah Run & 25±16 & \textbf{151±38} & 105±54 & 91±27 \\
            Finger Spin & 7±4 & 600±100 & 436±106 & \textbf{735±44} \textbf{*}\\
            Cartpole Swingup & 98±21 & 257±41 & 387±51 & \textbf{407±81} \\
            Cup Catch & 111±31 & 518±256 & 664±48 & \textbf{816±70} \textbf{*}\\
            \bottomrule
            \end{tabular}
        }
    \end{minipage}
    \vspace{4pt}

    \centering
    \begin{minipage}[b]{0.48\linewidth}
        \centering
        \captionsetup{font=small, skip=2pt} 
        \caption*{e) \textbf{Color Video Easy}}
        \resizebox{\linewidth}{!}{%
            \begin{tabular}{ccccc}
            \toprule
              &  DrQ &  DrQ + Aug & SVEA & SADA \\
            \midrule
            Walker Walk & 208±49 & 879±42 & 817±140 & \textbf{935±24} \\
            Walker Stand & 487±28 & 963±6 & 811±341 & \textbf{970±4} \\
            Cheetah Run & 60±36 & 263±48 & 294±27 & \textbf{331±57} \\
            Finger Spin & 310±30 & 920±42 & 866±137 & \textbf{972±4} \\
            Cartpole Swingup & 327±43 & 528±154 & \textbf{761±44} & 748±64 \\
            Cup Catch & 447±61 & 697±353 & 944±17 & \textbf{959±8} \\
            \bottomrule
            \end{tabular}
        }
    \end{minipage}
    \hspace{0.02\linewidth} 
    \begin{minipage}[b]{0.48\linewidth}
        \centering
        \captionsetup{font=small, skip=2pt} 
        \caption*{f) \textbf{Color Video Hard}}
        \resizebox{\linewidth}{!}{%
            \begin{tabular}{ccccc}
            \toprule
              &  DrQ &  DrQ + Aug & SVEA & SADA \\
            \midrule
            Walker Walk & 42±10 & 639±69 & 600±150 & \textbf{736±68} \\
            Walker Stand & 170±17 & 889±48 & 730±315 & \textbf{920±22} \\
            Cheetah Run & 26±17 & \textbf{216±53} & 153±66 & 187±63 \\
            Finger Spin & 2±2 & 684±82 & 500±151 & \textbf{815±25} \textbf{*}\\
            Cartpole Swingup & 94±17 & 300±57 & 464±63 & \textbf{469±80} \\
            Cup Catch & 122±48 & 570±300 & 792±63 & \textbf{873±43} \textbf{*}\\
            \bottomrule
            \end{tabular}
        }
    \end{minipage}

    \caption{\textbf{DMC-GB2 Photometric Test Set Results.} Episode Reward. Methods trained under photometric augmentations and evaluated on DMC-GB2 Photometric Test Set. Mean and Stddev over 5 random seeds. Highest scores in bold.  Asterisk (*) indicates that the method is statistically significantly greater than all compared methods with 95\% confidence.}
    \label{}

\end{figure}

\clearpage

\begin{figure}[H]
    \centering
    \includegraphics[width=1.0\linewidth]{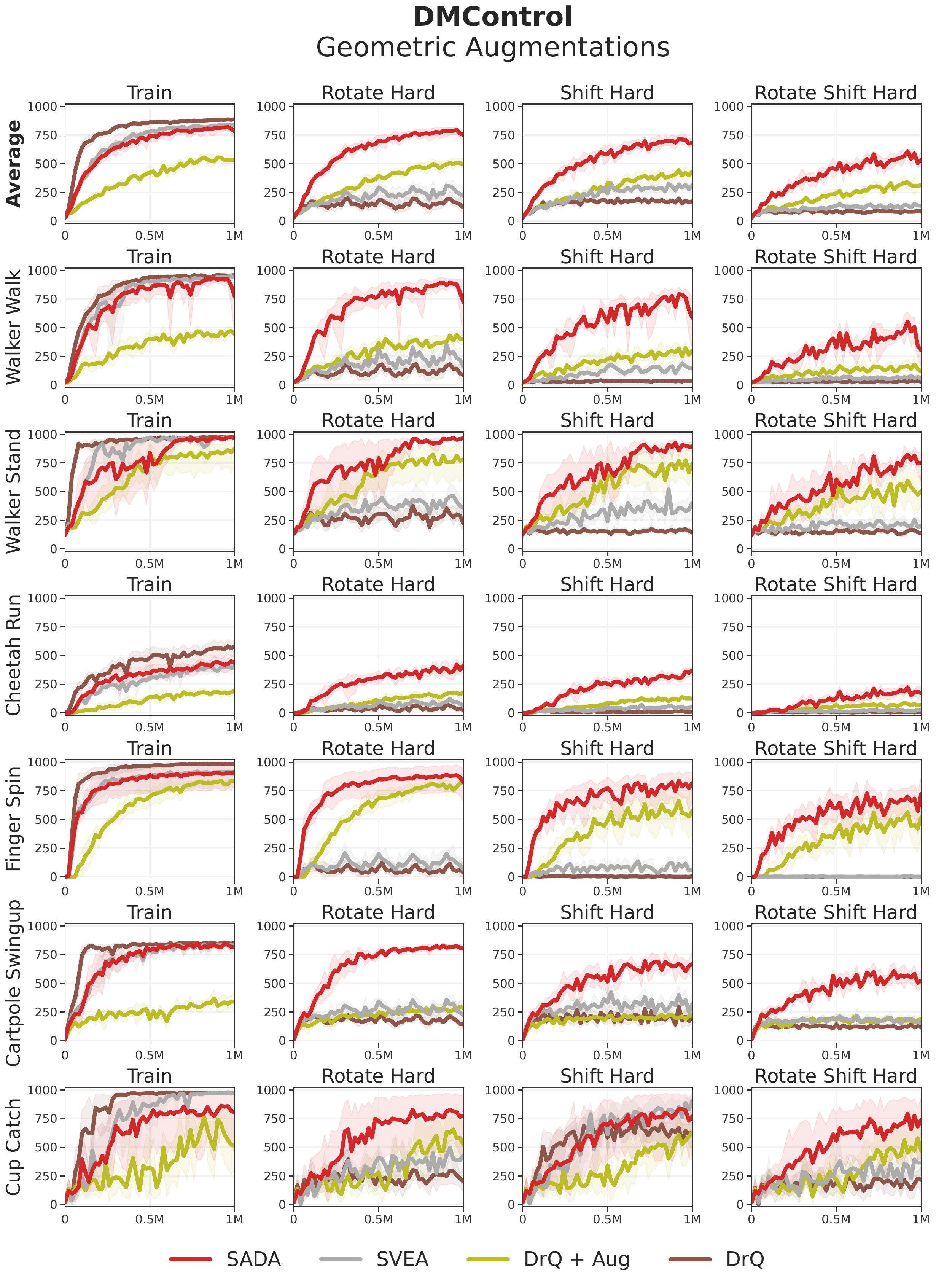}
    \caption{\textbf{DMC-GB2 Geometric Test Set Graphs.} Episode Reward. Methods trained under geometric augmentations and evaluated on DMC-GB2 Geometric Test Set. Hard levels visualized. Mean and 95\% CI over 5 random seeds.}
    \label{fig:app_dmc_geo_graphs}
\end{figure}

\begin{figure}[H]
    \centering
    \includegraphics[width=1.0\linewidth]{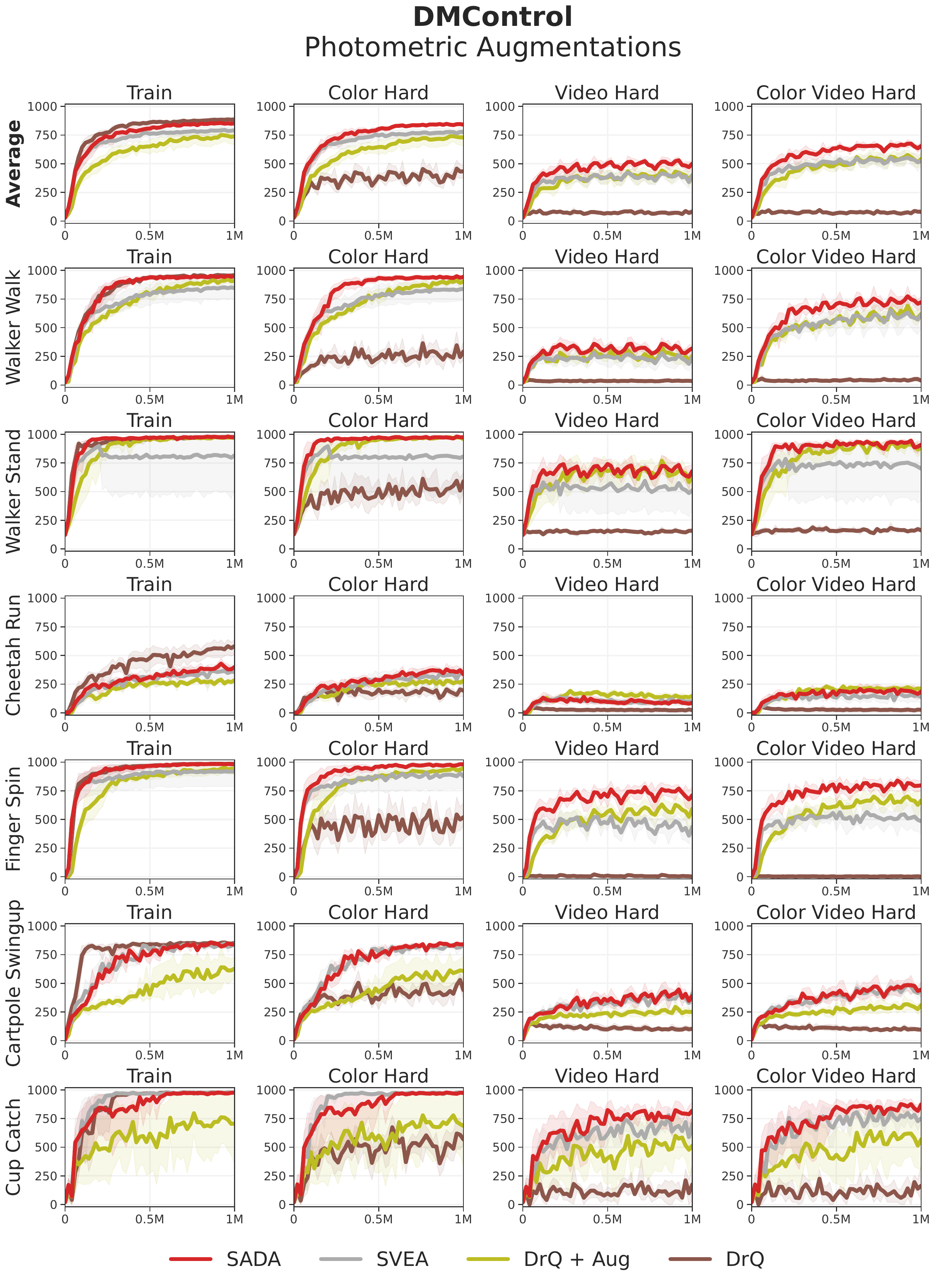}
    \caption{\textbf{DMC-GB2 Photometric Test Set Graphs.} Episode Reward. Methods trained under photometric augmentations and evaluated on DMC-GB2 Photometric Test Set. Hard levels visualized. Mean and 95\% CI over 5 random seeds.}
    \label{fig:app_dmc_photo_graphs}
\end{figure}

\clearpage
\subsection{Meta-World Results}
\label{subsection:app_mw_extended_results}

\begin{figure}[htbp]
    \centering
    \begin{minipage}[b]{0.48\linewidth}
        \centering
        \captionsetup{font=small, skip=2pt} 
        \caption*{\textbf{Shift Hard (Meta-World)}}
        \resizebox{\linewidth}{!}{%
            \begin{tabular}{ccccc}
            \toprule
             &  DrQ &  DrQ + Aug & SVEA & SADA \\
            \midrule
            Door Open & 2±2 & 51±12 & 28±7 & \textbf{59±9} \\
            Peg Unplug Side & 2±1 & 33±27 & 32±13 & \textbf{70±18} \textbf{*} \\
            Sweep Into & 3±2 &  \textbf{76±9} & 42±8 & 74±8 \\
            Basketball & 0±0 & 48±31 & 18±16 & \textbf{75±16} \\
            Push & 2±2 & 43±23 & 28±4 & \textbf{61±16} \\
            \bottomrule
            \end{tabular}
        }
    \end{minipage}
    \caption{\textbf{Meta-World Results.} Success rate (\%). Trained under strong shift augmentation only. Evaluated on Meta-World Shift Hard. Mean and Stddev of 5 random seeds. Highest scores in bold.  Asterisk (*) indicates that the method is statistically significantly greater than all compared methods with 95\% confidence.}

    \vspace{0.2in}
    
    \begin{minipage}[H]{0.5\linewidth}
        \centering
        \includegraphics[width=1.0\linewidth]{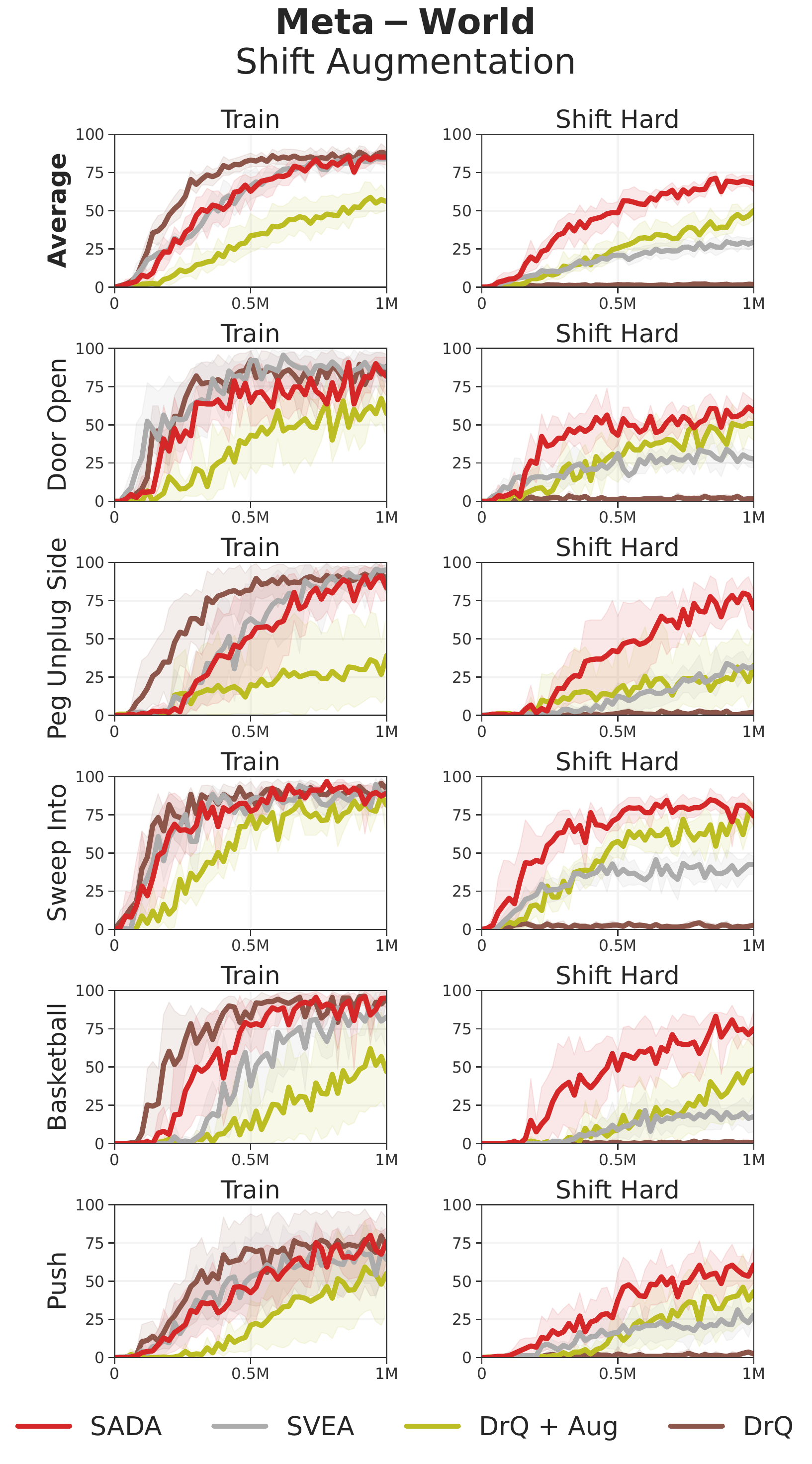}
    \end{minipage}
    \caption{\textbf{Meta-World Graphs.} Success rate (\%). Trained under Shift Augmentation, Evaluated on Meta-World Shift Hard. Mean and 95\% CI of 5 random seeds.}
    \label{fig:app_mw_extended_results}
\end{figure}

\clearpage
\section{Statistical Significance Testing}
\label{sec:stat_sig_test}
\vspace{-0.1cm}

We conduct statistical significance testing for all our experiments and provide it below. Given two methods $\mathcal{A}$ and $\mathcal{B}$, we use a one tailed Welch t-test to determine the statistical significance and formulate the following hypotheses:
\begin{align}
    \textbf{Null Hypothesis} ~~~\mathcal{H}_o\mathcal{: A\leq B}\\
    \textbf{Alternative Hypothesis} ~~~\mathcal{H}_a\mathcal{: A> B}
\end{align}

Using an alpha value of 0.05 (95\% confidence), all p-values greater than 0.05 indicate that the null hypothesis cannot be rejected and that the expected mean of $\mathcal{A}$ is statistically significantly \textbf{less than or equal to} the expected mean of $\mathcal{B}$. On the other hand, all p-values less than 0.05 indicate that we should reject the null hypothesis in favor of the alternative hypothesis, indicating that the expected mean of $\mathcal{A}$ is statistically significantly \textbf{greater than} the expected mean of $\mathcal{B}$. To control for multiple pairwise comparisons, we apply the Holm-Bonferroni method, where we sort the p-values in ascending order, and compare them with their adjusted alpha values (0.0167, 0.025, 0.05) respectively. Using the Holm-Bonferroni method, there is only a 5\% chance of rejecting at least one true null hypothesis (i.e., making a Type I error) from the three hypotheses in every comparison.

We provide per-task statistical significance testing results in the tables in Appendix \ref{sec:app_extended_results}. We also provide the overall category statistical significance testing results below. 

\subsection{Overall Category Results:}
\label{subsection:sig}
\textbf{For all the \emph{overall category} results of the experiments conducted throughout this paper, there is sufficient evidence (with 95\% confidence) that the mean performance of SADA is statistically significantly greater than all of the baselines.} 

In the overall category statistical significance testing, we provide both the \textbf{p} and \textbf{t} values for the Welch t-test results. \textbf{p} denotes the p-value which represents the probability of observing the data or more extreme data under the assumption that the null hypothesis is true. \textbf{t} denotes the test statistic which is a standardized measure of the difference between two group means, adjusted for the variability within the groups, used to assess the significance of the observed difference.

\begin{figure}[H]
    \centering
    \begin{minipage}[b]{1.0\linewidth}
        \centering
        \captionsetup{font=normal, skip=2pt} 
        \caption*{\textbf{Overall Robustness}}
        \setlength{\extrarowheight}{0.05in} 
        \resizebox{\linewidth}{!}{%
        \begin{tabular}{|p{0.45in}|p{1.2in}|p{1.4in}|p{1.4in}|p{1.4in}|} 
            \hline
            \multicolumn{2}{|c|}{\multirow{2}{*}{\textbf{Method} $\mathcal{A}$}} & \multicolumn{3}{c|}{\textbf{Method} $\mathcal{B}$}  \\ \cline{3-5} 
            \multicolumn{2}{|c|}{}   & \centering\textbf{SVEA} & \centering\textbf{DrQ + Aug} & \multicolumn{1}{>{\centering\arraybackslash}c|}{\textbf{DrQ}} \\ \cline{1-5} 
            \multirow{3}{*}{\textbf{SADA}} & \centering \textbf{Avg Geometric} & p=4.0$\times10^{-9}$, t=27.21 & p=8.1$\times10^{-10}$, t=30.62 & p=4.0$\times10^{-8}$, t=45.78  \\ \cline{2-5} 
             & \centering \textbf{Avg Photometric} & p=1.2$\times10^{-3}$, t=5.77 & p=1.4$\times10^{-10}$, t=47.60 & p=2.9$\times10^{-10}$, t=36.17 \\ \cline{2-5} 
             & \centering \textbf{Avg All} & p=3.4$\times10^{-6}$, t=15.42 & p=1.1$\times10^{-10}$, t=38.84 & p=1.9$\times10^{-9}$, t=44.27 \\ \hline 
        \end{tabular}
        }
    \end{minipage}
    \vspace{0.2cm}
    \caption{\textbf{Overall Robustness.} Statistical Significance Measurement using Welch t-test on the episode reward on DMC-GB2. Methods trained under all augmentations and averaged across all DMControl tasks. Mean over 5 random seeds.}
\end{figure}

\begin{figure}[H]
    \centering
    \begin{minipage}[b]{1.0\linewidth}
        \centering
        \captionsetup{font=normal, skip=2pt} 
        \caption*{\textbf{Geometric vs Photometric Robustness}}
        \setlength{\extrarowheight}{0.05in} 
        \resizebox{\linewidth}{!}{%
        \begin{tabular}{|p{0.45in}|p{1.2in}|p{1.4in}|p{1.4in}|p{1.4in}|} 
            \hline
            \multicolumn{2}{|c|}{\multirow{2}{*}{\textbf{Method} $\mathcal{A}$}} & \multicolumn{3}{c|}{\textbf{Method} $\mathcal{B}$}  \\ \cline{3-5} 
            \multicolumn{2}{|c|}{}   & \centering\textbf{SVEA} & \centering\textbf{DrQ + Aug} & \multicolumn{1}{>{\centering\arraybackslash}c|}{\textbf{DrQ}} \\ \cline{1-5} 
            \multirow{2}{*}{\textbf{SADA}} & \centering \textbf{Avg Geometric} & p=6.2$\times10^{-5}$, t=12.96 & p=8.8$\times10^{-5}$, t=7.53 & p=2.0$\times10^{-5}$, t=18.28 \\ \cline{2-5} 
             & \centering \textbf{Avg Photometric} & p=7.3$\times10^{-3}$, t=3.80 & p=1.4$\times10^{-2}$, t=3.29 & p=1.3$\times10^{-11}$, t=50.40 \\ \hline
        \end{tabular}
        }
    \end{minipage}
    \vspace{0.2cm}
    \caption{\textbf{Geometric vs Photometric Robustness.} Statistical Significance Measurement using Welch t-test on the episode reward on DMC-GB2. Methods were trained under geometric augmentations and evaluated on the geometric test set, and trained under photometric augmentations and evaluated on the photometric test set, averaged across all DMControl tasks. Mean over 5 random seeds.}
\end{figure}

\begin{figure}[H]
    \centering
    \begin{minipage}[b]{1.0\linewidth}
        \centering
        \captionsetup{font=normal, skip=2pt} 
        \caption*{\textbf{Ablations}}
        \setlength{\extrarowheight}{0.05in} 
        \resizebox{\linewidth}{!}{%
        \begin{tabular}{|p{0.45in}|p{1.2in}|p{1.4in}|p{1.4in}|p{1.4in}|} 
            \hline
            \multicolumn{2}{|c|}{\multirow{2}{*}{\textbf{Method} $\mathcal{A}$}} & \multicolumn{3}{c|}{\textbf{Method} $\mathcal{B}$}  \\ \cline{3-5} 
            \multicolumn{2}{|c|}{}   & \centering\textbf{SADA (Naive Actor Aug)} & \centering\textbf{SADA (Naive Critic Aug)} & \textbf{SADA (No Critic Aug)} \\ \cline{1-5} 
            \multirow{3}{*}{\textbf{SADA}} & \centering \textbf{Avg Geometric} & p=1.3$\times10^{-7}$, t=16.88 & p=6.3$\times10^{-9}$, t=23.43 & p=1.6$\times10^{-2}$, t=2.61 \\ \cline{2-5} 
             & \centering \textbf{Avg Photometric} & p=3.2$\times10^{-3}$, t=4.30 & p=2.0$\times10^{-7}$, t=22.86 & p=2.3$\times10^{-8}$, t=20.58 \\ \cline{2-5} 
             & \centering \textbf{Avg All} & p=1.1$\times10^{-5}$, t=10.89 & p=1.6$\times10^{-8}$, t=24.05 & p=1.3$\times10^{-6}$, t=12.01 \\ \hline 
        \end{tabular}
        }
    \end{minipage}
    \vspace{0.2cm}
    \caption{\textbf{Ablations.} Statistical Significance Measurement using Welch t-test on the episode reward on DMC-GB2. Methods trained under all augmentations and averaged across all DMControl tasks. Mean over 5 random seeds.}
\end{figure}

\begin{figure}[H]
    \centering
    \begin{minipage}[b]{0.7\linewidth}
        \centering
        \captionsetup{font=normal, skip=2pt} 
        \caption*{\textbf{TD-MPC2 Baseline}}
        \setlength{\extrarowheight}{0.05in} 
        \resizebox{\linewidth}{!}{%
        \begin{tabular}{|p{1.4in}|p{0.7in}|p{1.1in}|p{1.1in}|p{1.1in}|} 
            \hline
            \multicolumn{2}{|c|}{\multirow{2}{*}{\textbf{Method} $\mathcal{A}$}} & \multicolumn{3}{c|}{\textbf{Method} $\mathcal{B}$}  \\ \cline{3-5} 
            \multicolumn{2}{|c|}{}   &
            \multicolumn{3}{|c|}{\textbf{TD-MPC2 + Aug}} \\ \cline{1-5} 
            \textbf{TD-MPC2 + SADA} & \centering \textbf{Avg All} & \multicolumn{3}{|c|}{p=8.1$\times10^{-5}$, t=7.90 } \\ \hline
        \end{tabular}
        }
    \end{minipage}
    \vspace{0.2cm}
    \caption{\textbf{TD-MPC2 Baseline.} Statistical Significance Measurement using Welch t-test on the episode reward on DMC-GB2. Trained under all
    augmentations with a TD-MPC2 backbone, averaged across all DMControl tasks. Mean over 5 random seeds.}
\end{figure}

\begin{figure}[H]
    \centering
    \begin{minipage}[b]{1.0\linewidth}
        \centering
        \captionsetup{font=normal, skip=2pt} 
        \caption*{\textbf{Meta-World}}
        \setlength{\extrarowheight}{0.05in} 
        \resizebox{\linewidth}{!}{%
        \begin{tabular}{|p{0.45in}|p{1.2in}|p{1.4in}|p{1.4in}|p{1.4in}|} 
            \hline
            \multicolumn{2}{|c|}{\multirow{2}{*}{\textbf{Method} $\mathcal{A}$}} & \multicolumn{3}{c|}{\textbf{Method} $\mathcal{B}$}  \\ \cline{3-5} 
            \multicolumn{2}{|c|}{}   & \centering\textbf{SVEA} & \centering\textbf{DrQ + Aug} & \multicolumn{1}{>{\centering\arraybackslash}c|}{\textbf{DrQ}} \\ \cline{1-5} 
            \textbf{SADA} & \centering \textbf{Shift Hard} & p=1.5$\times10^{-5}$, t=10.26 & p=2.2$\times10^{-3}$, t=3.90 & p=1.4$\times10^{-5}$, t=20.45  \\ \hline
        \end{tabular}
        }
    \end{minipage}
    \vspace{0.2cm}
    \caption{\textbf{Meta-World.} Statistical Significance Measurement using Welch t-test on the success rate (\%) on Shift Hard (Meta-World) distribution. Trained under strong shift augmentation only, averaged across all Meta-World tasks. Mean over 5 random seeds.}
\end{figure}

\end{document}